\pdfoutput=1
\documentclass{article}

\usepackage{preprint}

\usepackage{times}
\usepackage{amsmath,amssymb,amsfonts}
\usepackage{mathtools}
\usepackage{graphicx}
\usepackage{tikz}
\usetikzlibrary{arrows.meta,decorations.pathreplacing,calc}
\usepackage{booktabs}
\usepackage{multirow}
\usepackage{enumitem}
\usepackage{wrapfig}

\renewenvironment{wrapfigure}[3][]{%
  \figure[t]\centering
  \tikzset{every picture/.append style={scale=1.45,transform shape}}%
}{\endfigure}

\usepackage{xspace}
\usepackage{xcolor}
\usepackage{hyperref}
\usepackage{url}
\hypersetup{colorlinks=true, linkcolor=blue!55!black, citecolor=blue!55!black, urlcolor=blue!55!black}

\newcommand{\method}{LiLiCorr\xspace}
\newcommand{\dflash}{DFlash\xspace}
\newcommand{\methodc}{LiLiCorr\,+\,conv\xspace}
\newcommand{\dflashtwo}{DFlash2\xspace}

\DeclareMathOperator*{\argmax}{arg\,max}

\DeclareMathOperator{\softmax}{softmax}
\DeclareMathOperator{\cossim}{cos}
\DeclareMathOperator{\relu}{ReLU}

\newcommand{\vc}[1]{\mathbf{#1}}
\newcommand{\reals}{\mathbb{R}}
\newcommand{\given}{\mid}

\newcommand{\Epair}{\phi^{\mathrm{p}}}
\newcommand{\outvec}{\vc{o}}
\newcommand{\invec}{\vc{i}}

\newcommand{\topk}{K}
\newcommand{\block}{B}

\providecommand{\authornote}{}

\title{\method: Lightweight Likelihood Correlation\\of Parallel Drafts for Speculative Decoding}
\shorttitle{\method: Lightweight Likelihood Correlation of Parallel Drafts for Speculative Decoding}

\newcommand{\upd}[1]{}      
\newcommand{\chg}[1]{#1}    

\author{%
Matan Rusanovsky\thanks{Correspondence to \texttt{mrusanovsky@nvidia.com}.}
\quad Yoav Miron \quad Roy Uziel \quad Omer Belhasin \quad Hao Guo\\[3pt]
Ran Zilberstein \quad Maor Ashkenazi \quad Michael Elad\\[5pt]
NVIDIA
}

\begin{document}
\maketitle
\authornote

\begin{abstract}
Speculative decoding accelerates language-model inference by drafting future
tokens that the target model verifies in parallel. A diffusion-style drafter, such as \dflash,
drafts an entire block in one forward pass. It is trained on the per-position marginals rather than
on the joint distribution over the block, so the tokens it emits are individually plausible yet
jointly incoherent. We introduce \method, a {\bf Li}ghtweight {\bf Li}kelihood-based model that
{\bf Corr}elates the per-position marginal distributions such a drafter produces.
It keeps the top-$\topk$ tokens at each position as \emph{candidates} and processes them
jointly, producing for each an \emph{in} and an \emph{out} vector. A pair at consecutive positions matches
when the earlier candidate's \emph{out} vector aligns, in cosine similarity, with the later one's
\emph{in} vector. Training makes the correct pairings score highest while pushing
competing ones down, so coherent blocks outscore incoherent ones. The joint distribution over the
block, exponential in its length, is never materialized. One lightweight network
pass produces all the vectors, and the pairwise scores follow as batched matrix operations,
leaving only a cheap greedy walk sequential. We
co-train the \dflash drafter with \method, so it learns to propose
candidates that correlate into longer accepted sequences. \chg{Over the vanilla \dflash
drafter it builds on, \method accepts more and serves faster at every one of the $72$ settings we
test: nine benchmarks at two target sizes under greedy and temperature-one decoding, together
with a throughput sweep over six concurrencies, two input lengths and three output-entropy tiers.
It raises acceptance length by $7$ to $19\%$, while its
single-pass scoring head accounts for only about $3\%$ of the per-block latency. Against
three concurrently developed methods that also restore coherence at draft time, every system
equally optimized on a common serving stack, \method holds the highest
throughput in $63$ of those settings, ties within a measured noise floor in $6$, and trails in
only $3$.}

\end{abstract}

\section{Introduction}

Speculative decoding accelerates autoregressive language-model inference by letting a
small draft model propose several future tokens that the target model verifies in
parallel, keeping the longest correct prefix \citep{leviathan2023speculative,chen2023accelerating}.
Because decoding is memory-bandwidth bound, verifying a multitude of tokens together costs
little more than generating just one token. The resulting speedup depends on two quantities: the
\emph{acceptance length}, the expected number of draft tokens accepted per step, and
the cost of the drafter itself. A good drafter must raise the former without inflating
the latter. Much effort therefore goes into raising acceptance length, with designs
ranging from feature-autoregressive heads such as EAGLE
\citep{li2024eagle,li2024eagle2,li2025eagle3} to multi-head predictors such as Medusa
\citep{cai2024medusa}. A particularly promising approach predicts an entire block of
future tokens in a single forward pass via a parallel diffusion-style drafter,
such as \dflash \citep{dflash2025},  offering a long speculative window at very low
cost.

In this work we build on \dflash, which predicts an entire block of future tokens in one forward
pass. For each position in the block, henceforth a \emph{slot}, it produces a
distribution over the full vocabulary. Although \dflash attends bidirectionally across
slots, so its representations can in principle correlate different positions, it is
optimized with a per-position cross-entropy that supervises only the token
\emph{marginal} at each slot. This leaves the \emph{joint} distribution over the block
largely unconstrained, so \dflash predicts each slot well yet can assemble the tokens into an
incoherent block.

This marginals problem is well known in the discrete diffusion literature, and its
origin is combinatorial: the joint distribution over a block of $\block$ positions, each
drawn from a vocabulary of size $V$, has $V^{\block}$ possible realizations, so
materializing it is intractable even for very short blocks. Diffusion language models
therefore never model the joint explicitly, and both diffusion and speculative-decoding literature offer ways to recover coherence without it. Masked
diffusion models such as LLaDA, MDLM, and Dream
\citep{nie2025llada,sahoo2024mdlm,dream2025} correlate the sequence through iterative
denoising, re-masking a subset of the predicted tokens and re-predicting them over
several rounds. In a speculative-decoding setting, however, each such round is another
full drafter pass, which is too expensive. Tree-based reranking, such as DDTree
\citep{ddtree}, expands several candidate paths from the per-slot marginals and lets
the target model select among them. This shifts the correlation work onto the target,
whose passes speculative decoding is designed to save. The trade is especially
unfavorable in the compute-bound, high-concurrency regime typical of real-world
serving. 
In contrast, Domino \citep{domino2026} and DSpark \citep{dspark2026}, \chg{two works concurrent
with ours}, act at draft time while avoiding extra target passes, but
they do so through a sequence of auto-regressive neural-network passes, one on each draft slot.
\chg{A third work developed concurrently with ours, \dflashtwo \citep{dflash2_2026}, avoids the
per-slot pass, scoring each adjacent pair of tokens from learned embeddings of their two identities.
Compatibility is therefore a property of the token pair itself rather than of the draft it appears
in: the same two tokens score alike in every context.}

\definecolor{outcol}{HTML}{C2410C}
\definecolor{incol}{HTML}{125CA6}
\definecolor{okgreen}{HTML}{1B7837}
\definecolor{badred}{HTML}{A32020}

\providecommand{\lnkfaint}[4]{%
  \draw[outcol,opacity=0.30,line width=0.5pt] (#1,#2) -- ($(#1,#2)!0.5!(#3,#4)$);
  \draw[incol,opacity=0.30,line width=0.5pt] ($(#1,#2)!0.5!(#3,#4)$) -- (#3,#4);}
\providecommand{\lnkbold}[4]{%
  \draw[outcol,line width=1.4pt] (#1,#2) -- ($(#1,#2)!0.5!(#3,#4)$);
  \draw[incol,line width=1.4pt] ($(#1,#2)!0.5!(#3,#4)$) -- (#3,#4);}

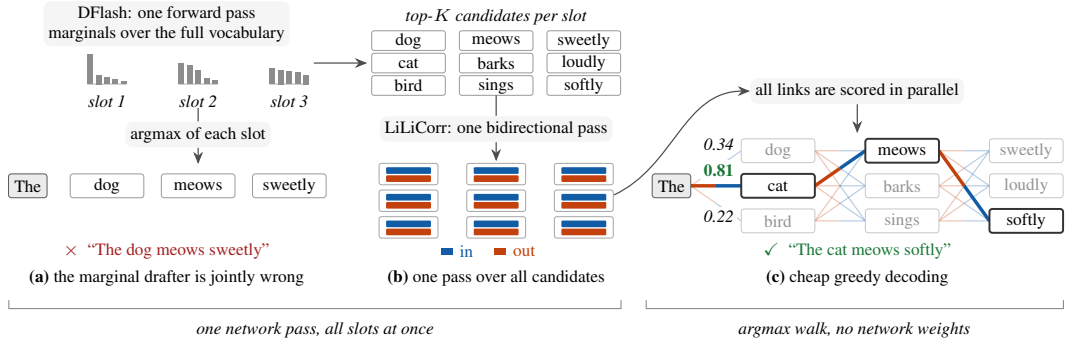
\begin{figure}[t]
\centering
\resizebox{\textwidth}{!}{%
\begin{tikzpicture}[x=1cm,y=1cm,font=\scriptsize,
  chip/.style={draw=black!40,line width=0.4pt,fill=white,rounded corners=1.2pt,
               minimum width=1.05cm,minimum height=0.34cm,inner sep=1pt},
  anch/.style={chip,minimum width=0.55cm,fill=black!8,draw=black!65},
  dim/.style={chip,draw=black!22,text=black!42},
  pick/.style={chip,draw=black!80,line width=0.8pt},
  panel/.style={fill=black!4,rounded corners=3pt,inner sep=2pt,align=center},
  note/.style={inner sep=1pt,font=\scriptsize\itshape},
  lb/.style={inner sep=1pt},
  ar/.style={-{Stealth[length=5pt,width=4pt]},draw=black!70,line width=0.5pt},
  shaft/.style={draw=black!70,line width=0.5pt}]

\node[panel] (dfb) at (2.30,4.86) {\dflash: one forward pass\\ marginals over the full vocabulary};

\foreach \cx in {1.45,2.75,4.05}
  \draw[black!30,line width=0.3pt] ({\cx-0.28},4.00) -- ({\cx+0.28},4.00);
\foreach \x/\h in {-0.24/0.42,-0.12/0.13,0.00/0.09,0.12/0.06,0.24/0.04}
  \fill[black!45] ({1.45+\x-0.04},4.00) rectangle ++(0.08,\h);
\foreach \x/\h in {-0.24/0.29,-0.12/0.27,0.00/0.19,0.12/0.08,0.24/0.05}
  \fill[black!45] ({2.75+\x-0.04},4.00) rectangle ++(0.08,\h);
\foreach \x/\h in {-0.24/0.22,-0.12/0.20,0.00/0.18,0.12/0.16,0.24/0.13}
  \fill[black!45] ({4.05+\x-0.04},4.00) rectangle ++(0.08,\h);
\node[note] at (1.45,3.76) {slot 1};
\node[note] at (2.75,3.76) {slot 2};
\node[note] at (4.05,3.76) {slot 3};

\node[panel] (amx) at (2.75,3.28) {argmax of each slot};
\draw[shaft] (2.75,3.66) -- (amx.north);
\draw[ar] (amx.south) -- (2.75,2.86);

\node[anch] at (0.32,2.55) {The};
\node[chip] at (1.45,2.55) {dog};
\node[chip] at (2.75,2.55) {meows};
\node[chip] at (4.05,2.55) {sweetly};
\node[lb,text=badred] at (2.30,1.62) {$\times$~~``The dog meows sweetly''};

\draw[ar] (4.40,4.30) -- (5.16,4.30);
\node[note] at (7.00,4.96) {top-$\topk$ candidates per slot};

\foreach \cx/\ta/\tb/\tc in {5.75/dog/cat/bird, 7.00/meows/barks/sings,
                             8.25/sweetly/loudly/softly} {
  \foreach \cy/\tk in {4.62/\ta, 4.30/\tb, 3.98/\tc} {
    \draw[black!40,line width=0.4pt,fill=white,rounded corners=1pt]
      ({\cx-0.525},{\cy-0.14}) rectangle ({\cx+0.525},{\cy+0.14});
    \node[lb] at (\cx,\cy) {\tk};
  }
}
\node[panel] (lcb) at (7.00,3.36) {\method: one bidirectional pass};
\draw[shaft] (7.00,3.84) -- (lcb.north);
\draw[ar] (lcb.south) -- (7.00,2.90);

\foreach \cx in {5.75,7.00,8.25} {
  \foreach \cy in {2.72,2.34,1.96} {
    \draw[black!35,line width=0.4pt,fill=white,rounded corners=1pt]
      ({\cx-0.42},{\cy-0.155}) rectangle ({\cx+0.42},{\cy+0.15});
    \fill[incol,rounded corners=0.5pt] ({\cx-0.32},\cy) rectangle ({\cx+0.32},{\cy+0.095});
    \fill[outcol,rounded corners=0.5pt] ({\cx-0.32},{\cy-0.115}) rectangle ({\cx+0.32},{\cy-0.02});
  }
}
\fill[incol] (6.22,1.575) rectangle (6.38,1.665);
\node[lb,anchor=west,text=incol] at (6.44,1.62) {in};
\fill[outcol] (6.98,1.575) rectangle (7.14,1.665);
\node[lb,anchor=west,text=outcol] at (7.20,1.62) {out};


\foreach \yb in {3.05,2.55,2.05} {\lnkfaint{9.775}{2.55}{10.495}{\yb}}
\foreach \ya in {3.05,2.55,2.05} {\foreach \yb in {3.05,2.55,2.05}
  {\lnkfaint{11.545}{\ya}{12.265}{\yb}}}
\foreach \ya in {3.05,2.55,2.05} {\foreach \yb in {3.05,2.55,2.05}
  {\lnkfaint{13.315}{\ya}{14.035}{\yb}}}
\lnkbold{9.775}{2.55}{10.495}{2.55}
\lnkbold{11.545}{2.55}{12.265}{3.05}
\lnkbold{13.315}{3.05}{14.035}{2.05}

\node[note,fill=white,inner sep=1.2pt,anchor=east] at (10.44,3.14) {0.34};
\node[lb,fill=white,inner sep=1.2pt,anchor=east,text=okgreen] at (10.44,2.78)
  {\textbf{0.81}};
\node[note,fill=white,inner sep=1.2pt,anchor=east] at (10.44,2.12) {0.22};

\node[anch] at (9.50,2.55) {The};
\node[dim] at (11.02,3.05) {dog};    \node[pick] at (11.02,2.55) {cat};
\node[dim] at (11.02,2.05) {bird};
\node[pick] at (12.79,3.05) {meows}; \node[dim] at (12.79,2.55) {barks};
\node[dim] at (12.79,2.05) {sings};
\node[dim] at (14.56,3.05) {sweetly}; \node[dim] at (14.56,2.55) {loudly};
\node[pick] at (14.56,2.05) {softly};

\node[panel] (alp) at (12.15,3.90) {all links are scored in parallel};
\draw[ar] (8.68,2.42) to[out=25,in=180] (alp.west);
\draw[ar] (alp.south) -- (12.15,3.30);
\node[lb,text=okgreen] at (12.15,1.62) {$\checkmark$~~``The cat meows softly''};

\draw[black!45,line width=0.45pt] (0.05,0.92) -- (0.05,0.80) -- (8.78,0.80) -- (8.78,0.92);
\draw[black!45,line width=0.45pt] (9.14,0.92) -- (9.14,0.80) -- (15.09,0.80) -- (15.09,0.92);
\node[note] at (4.42,0.50) {one network pass, all slots at once};
\node[note] at (12.11,0.50) {argmax walk, no network weights};

\node[lb] at (2.30,1.24) {\textbf{(a)} the marginal drafter is jointly wrong};
\node[lb] at (7.00,1.24) {\textbf{(b)} one pass over all candidates};
\node[lb] at (12.15,1.24) {\textbf{(c)} cheap greedy decoding};

\end{tikzpicture}}
\caption{Overview of \method. \textbf{(a)} \dflash produces a distribution logit over the vocabulary
at each slot in one forward pass, and taking the most probable token of each slot independently
yields a sequence that is locally probable but jointly incoherent. \textbf{(b)} \method retains the
top-$\topk$ tokens of every slot and scores them jointly in a single pass, emitting an \emph{in} and
an \emph{out} vector per candidate. \textbf{(c)} Adjacent candidates are coupled by the alignment of
the earlier \emph{out} vector with the later \emph{in} vector, so each slot pair reduces to one
matrix product and the whole block is scored in parallel. The block is then committed left to right
by a greedy argmax operation that touches no network weights.}
\label{fig:teaser}
\end{figure}

We introduce \method, a lightweight likelihood-based model that recovers a coherent
sequence from the per-slot marginals in a \emph{single} network pass. Figure~\ref{fig:teaser} sets out the method end to end. \method
takes the top-$\topk$ candidate tokens at each slot and produces two vectors per
candidate: \emph{in} and \emph{out}. These capture how candidates chain together: two
adjacent candidates are compatible when the \emph{out} vector of the first aligns, in
cosine similarity, with the \emph{in} vector of the second. Training aligns the \emph{out}
and \emph{in} vectors of consecutive ground-truth tokens and separates the competing
pairings. At inference, a cheap greedy walk commits the block left to right,
at each position taking the candidate that best matches its committed predecessor. \chg{The novelty is that
every vector is produced in a single pass over all candidates at all slots, so each pair is scored
in the light of the candidates around it, at the slots before it and after it. Correlation then
reduces to} \emph{matrix operations that are
easy to optimize}\chg{, all candidate interactions computed in parallel, with only the greedy path
selection left sequential. Approaches that correct one slot at a time instead run their network at
every position.}

\method is trained jointly with \dflash rather than on top of a frozen drafter, so the
drafter learns to propose candidate sets that \method can turn into longer accepted
sequences. Against a vanilla \dflash drafter trained on the same data, this raises
acceptance length \chg{by $7$ to $19\%$} on every benchmark, for 8B and 4B target sizes and both greedy and
temperature-$1$ decoding. We \chg{also} compare against the \chg{three works above, Domino, DSpark
and \dflashtwo}, which likewise tune the drafter alongside their head, giving
every system the same optimization effort on a common serving stack.
\chg{\method is the fastest system at almost every setting we measure, and most of the few
exceptions sit inside the measurement noise. Against the two systems that correct one slot at a
time the advantage is structural, since they spend a network pass per slot where \method scores
every candidate at once. Against \dflashtwo, which also drafts in one pass, the difference lies in the
scoring rule. \dflashtwo scores a pair from the identities of its two tokens, so the same pair
scores alike in every draft, whereas every \method score is formed from vectors that have seen
every candidate at every slot. This structural difference yields longer accepted prefixes, on
average at both target sizes and under both decoding regimes. The lead persists as
concurrency rises.}

\paragraph{Contributions.}
\begin{enumerate}[label=(\arabic*),leftmargin=1.8em,topsep=1pt,itemsep=1pt,parsep=0pt]
\item \textbf{A novel likelihood-based formulation of the marginals problem.} We cast the recovery of a coherent sequence from per-slot marginals as a learned scoring model over candidate paths. Rather than reason in the full vocabulary, the model attends over all candidates of all slots by assigning two low-dimensional vectors to each, which encode how neighbors are matched. 
\item \textbf{Correlation as efficient parallel matrix operations.} At decoding time, a single
network pass precomputes 
all of the correlations, as the pairwise candidate interactions
reduce to 
matrix operations that are easy to optimize. This is followed by a cheap greedy walk that remains sequential.
\item \textbf{State-of-the-art results.} \method sets a new state-of-the-art for
draft-time correlation in speculative decoding, trained jointly with the drafter and given the same
data, training budget, and optimization effort as the \chg{three} concurrent
works~\chg{\cite{domino2026,dspark2026,dflash2_2026}}. We evaluate on nine benchmarks, at two target sizes and under
both greedy and temperature-one decoding, and on a throughput sweep over six serving concurrencies,
two input lengths, and three output-entropy tiers. \chg{\method holds the highest throughput of any
system compared in $63$ of these $72$ settings, ties within a measured noise floor in $6$ and
trails in only $3$.}
\end{enumerate}

\section{Related Work}
\label{sec:related}

\paragraph{Speculative decoding and parallel drafters.} Speculative decoding
accelerates inference by proposing draft tokens and verifying them with the target
model under a rule that preserves its distribution
\citep{leviathan2023speculative,chen2023accelerating}. Most work along these lines improves the
drafter, from multi-head predictors such as Medusa \citep{cai2024medusa} to
feature-autoregressive heads such as the EAGLE family
\citep{li2024eagle,li2024eagle2,li2025eagle3}. A separate line of work drafts an entire block in one
forward pass: parallel and diffusion-style block heads such as \dflash
\citep{dflash2025} and diffusion-based self-speculation \citep{nld2026} handle a long
speculative window cheaply. D-PACE \citep{dpace2026} improves \dflash with an
acceptance-aware training objective. Their parallel
prediction mirrors that of diffusion language models, such as LLaDA
\citep{nie2025llada}, MDLM \citep{sahoo2024mdlm}, Dream \citep{dream2025}, and
Nemotron-Labs-Diffusion \citep{nld2026}. These models exhibit a well-known marginals
limitation: although attention is bidirectional, the block is factorized into
independent per-position predictions, so the sampled tokens are not guaranteed to be
jointly coherent. This is addressed via an iterative refinement, re-masking and re-predicting
subsets of tokens so that later rounds condition on earlier ones. Parallel drafters
inherit the same limitation, but in speculative decoding each refinement round adds a
full drafter pass, which negates the latency gains that motivate parallel drafting. Recovering
coherence without repeated drafter passes is the problem \method addresses.

\paragraph{Recovering coherent drafts.} Several recent methods recover a coherent
sequence from the per-slot marginals that a parallel drafter such as \dflash produces.
One direction expands these
marginals into a draft tree of several continuations that the target verifies in a
single pass. DDTree \citep{ddtree} is training-free, ranking candidate prefixes by
their marginal product, whereas \citet{tfm2026} first learn a small autoregressive
adapter before building the tree. Both push the correlation work onto the target,
spending the compute speculative decoding aims to save, and both rely on
tree-structured attention, which is typically less efficient than standard causal
attention. An alternative direction reintroduces the dependencies at draft time: Domino
\citep{domino2026} and DSpark \citep{dspark2026}, \chg{two of the works developed concurrently with ours,}
condition each slot on the previously
sampled tokens, which avoids extra target passes but runs a correction network once at
every slot.
\chg{A third work developed concurrently with ours, \dflashtwo \citep{dflash2_2026}, correlates the
block without such a per-slot pass. It learns two embedding tables over the vocabulary, one holding
a vector for each token as a predecessor and one for each token as a successor, and scores a pair of
neighboring tokens by the inner product of the two. The score is added to the drafter's logits, and
the block is filled left to right, each slot committing the candidate that scores best against the
token just committed. What a pair receives is therefore fixed by the identities of its two tokens,
the same score in every draft. \dflashtwo introduces a second and independent contribution, a grouped
causal convolution inside each draft layer that lets neighboring slots exchange local information
while the block stays parallel. The convolution acts on the drafter rather than on the correlation,
so the two compose, and we ablate it on top of \method in Appendix~\ref{app:conv}.}
\chg{The marginals problem is not unique to speculative decoding: non-autoregressive translation
meets it in the same form, and \citet{sun2019fast} correlate a parallel decoder by factorizing
adjacent-pair scores into per-token vectors over a top-$\topk$ pool. What differs here is that
speculative decoding credits only an accepted prefix, which is why \method normalizes each slot
locally and commits greedily rather than decoding the best-scoring path.}
\method also acts at draft time \chg{and in a single pass}: it produces
all candidate representations at once, computes their pairwise correlations in
parallel, and leaves only a cheap greedy argmax over the resulting scores to run
sequentially. \chg{Its vectors are formed from every candidate in the block, so the score of a pair
reflects the context of the entire draft rather than the two token identities alone.}


\section{Method}
\label{sec:method}

\subsection{Setup: prefix acceptance and the marginal drafter}
\label{sec:setup}

Let $M$ be a frozen target language model. A parallel block drafter proposes, in a
single forward pass, a block of $\block$ positions conditioned on the verified
context, of which $\block-1$ are future tokens. The first block position is the last token
the target has committed, which serves as the \emph{anchor}, slot $0$. The remaining $\block-1$ positions are
the speculative \emph{slots} $1, \dots, \block-1$. At each slot the drafter produces a distribution (logit) over the
vocabulary, a per-slot \emph{marginal}. Standard decoding resolves each slot
independently, taking its most probable token. Once the draft is complete, the
target verifies the whole block in a single pass and accepts a \emph{prefix} of it,
discarding everything from the first rejected position onward. With a greedy target a draft
token is accepted when it coincides with the target's own choice. Under a sampling target,
acceptance instead follows the standard rejection test, which preserves the target distribution
exactly \citep{leviathan2023speculative,chen2023accelerating}. In both regimes acceptance is
prefix-structured, and it is this structure that \method is built around.

\subsection{A scoring model over the candidate lattice}
\label{sec:lattice}

The drafter is supervised on per-slot marginals, so the tokens it selects are
individually probable but not necessarily compatible with one another. Selecting tokens
that are jointly compatible, and not only locally probable, can lengthen the accepted
prefix, and doing so at draft time could avoid extra target passes.
\method addresses this goal by correlating the slots over a small pool of alternatives that the
drafter's own marginal already places near the top. At each slot $s$ it retains the drafter's
top-$\topk$ candidate tokens $c_{s,1}, \dots, c_{s,\topk}$ rather than its single most probable
one. Each candidate is a token, and we represent it by a vector we call its \emph{node},
formed by summing four terms of equal width: the target's frozen input embedding of the token, the
drafter's hidden state at that slot, a small network over a few scalar features of how confident
the drafter is in that candidate, listed in Appendix~\ref{app:impl}, and learned encodings of the
slot and of the candidate's rank. The hidden state belongs to the slot and is shared by its $\topk$
candidates, so what separates them is the token itself, its confidence, and its rank. Across the
$\block-1$ slots these nodes form a $(\block\!-\!1) \times \topk$ lattice.

A small Transformer processes the $(\block\!-\!1) \times \topk$ nodes jointly. Its self-attention
lets every candidate exchange information with every other, both the competing candidates
within a slot and the candidates at neighboring slots, so each is represented in
the context of the whole block rather than on its own. Each attended node is then fused with a projection of the target's hidden state that produced
the last verified token: the candidate's own state, that projection, and their elementwise product
are concatenated and mapped back to the node width by a single layer. Though one vector common to
the whole block, it enters every candidate separately, so that the verifier's view of the context,
and not the drafter's alone, shapes each of them.
For each node \method emits two
vectors in $\reals^{d_e}$, an \emph{out} vector $\outvec$ and an \emph{in} vector $\invec$, each
normalized to unit norm. The dimension $d_e$ is given in Appendix~\ref{app:impl}. The anchor,
which precedes the first slot, is encoded far more thinly: the anchor projection itself, mapped by
its own head to one further out vector. Indexing the anchor as slot $0$, we
write this vector $\outvec_{0}$.
Adjacent slots $s$ and $s+1$ are then coupled by a single \emph{matching} term, the cosine
similarity between the out and in vectors:
\begin{equation}
\Epair(s,k,k') = \sigma \, \cossim(\outvec_{s,k},\, \invec_{s+1,k'}) ~~~~\mbox{for}~~~ s = 0, \dots, \block-2,~~~1\le k \le \topk_s,~1\le k' \le \topk,
\label{eq:pair}
\end{equation}
where $\cossim(\cdot,\cdot)$ is the cosine similarity and $\sigma$ is a fixed scaling
constant, with $\topk_0\!=\!1$ at the anchor, which carries a single out vector, and
$\topk_s\!=\!\topk$ at every speculative slot. Figure~\ref{fig:teaser}(b) shows the lattice and the pair of vectors each
candidate carries.

The factorization is local, but the information behind it is not. Only adjacent slots are
coupled by a score, yet the vectors that form that score depend on the entire lattice, so a pair at
slots $s$ and $s\!+\!1$ may be judged compatible on evidence lying outside them, among the
candidates that precede $s$ or follow $s\!+\!1$. We hypothesize that this is the more useful place
to spend capacity: rather than enrich the coupling itself, we keep it pairwise and let the vectors
it multiplies see the whole block. The pairwise form is what leaves the scores computable as
batched matrix products.

Casting each score as a cosine is a deliberate stabilizer: it removes vector magnitude as an
unconstrained degree of freedom. Without this bound we observe runaway score magnitudes that
corrupt, and in practice destroy, the shared \dflash backbone under joint training. The cosine
instead confines every learned factor to $[-\sigma, \sigma]$ while avoiding saturating
nonlinearities whose gradients vanish. We use a fixed scale $\sigma$ to separate the candidate scores to discriminate
among them. Its value, and the architecture of the head, are given in Appendix~\ref{app:impl}.

\subsection{Training the reranker}
\label{sec:objective}

\paragraph{Local normalization aligned with prefix acceptance.}
A natural formulation scores each full path by summing its matching terms across the block
and selects the path that maximizes this whole-chain score. We argue that this is the wrong target
\chg{under prefix acceptance}. The global optimum may sacrifice an early slot to win a richer tail, yet
under prefix acceptance that tail is never reached. Section~\ref{sec:decode} and
Appendix~\ref{app:decode} take up this comparison at decoding time. Optimizing the whole
chain thus spends capacity on regions the metric never credits. We therefore score each slot \emph{locally},
conditioned on the candidate committed just before it. Writing the per-slot
\emph{node score} of candidate $k$ as $z_s[k]$, the chain rule factorizes the path distribution
as
\begin{equation}
P_\theta(c_s \given c_{s-1}) = \softmax_k\big(z_s[k]\big),
\qquad z_s[k] = \Epair(s\!-\!1,\, c_{s-1},\, k),
\qquad s = 1, \dots, \block-1,
\label{eq:localar}
\end{equation}
where $c_s$ is the candidate committed at slot $s$ and $c_0$ is the anchor.
Each conditional still depends on the previous pick through the matching term in $z_s[k]$, so
locality avoids scoring the whole chain without reducing the model to a per-slot reranker.

\paragraph{Oracle-prefix supervision.} We train \method as a teacher-forced reranker over
the drafter's marginals, so the supervision must reflect how it is served. Two conditions
decide which slots carry a learning signal. The first is coverage: a slot is useful only when its
ground-truth token is among the drafter's $\topk$ candidates, since otherwise no reranking can
recover it. The second follows from prefix acceptance. We condition each transition on the
ground-truth predecessor $g_{s-1}$: a path that has already left the ground truth is discarded by
the verifier, so the transitions leaving it are never exercised in serving. A slot is therefore
supervised only when every earlier slot is covered as well. Together these define the
\emph{oracle prefix}, the longest leading run of slots that each contain the ground-truth token
among their $\topk$ candidates. Let $\mathcal{S}$ collect these slots across a training batch. We
teacher-force the per-slot cross-entropy on them:
\begin{equation}
\mathcal{L}_{\mathrm{CE}} \;=\; -\frac{1}{|\mathcal{S}|}\!\!\sum_{s \in \mathcal{S}} \log P_\theta\big(c_s = g_s \given c_{s-1} = g_{s-1}\big),
\label{eq:ce}
\end{equation}
where $g_s$ is the ground-truth token at slot $s$. The softmax of Eq.~\eqref{eq:localar} runs
over all $\topk$ candidates at slot $s$, so each supervised transition raises the ground-truth
successor and pushes its $\topk\!-\!1$ competitors down. A block whose very first slot already
misses contributes zero loss. The effect resembles the per-slot decay used to train \dflash,
though it arises from the coverage mask rather than from a designed weighting: supervision
concentrates on earlier slots.

\paragraph{Target-weighted distractor penalty.}
The per-slot cross-entropy raises the ground-truth probability but is indifferent to how the
residual mass is distributed among the remaining candidates. Under cross-entropy supervision
alone, we observed that \method's errors were typically candidates the target scored well-below the ground truth, a mismatch with the verifier rather than a genuine ambiguity in the
target distribution. Since the target is itself the verifier, its logits provide a graded measure
of candidate compatibility that cross-entropy ignores. Scoring against the target lets us direct
the negative gradient at these high-scoring but target-incompatible competitors, widening the
ground truth's margin over the candidates most likely to trigger an early verification failure.

Let $q_k$ denote the target probability of candidate $k$, and $g_s$ the ground-truth candidate at
slot $s$. Both $q_{g_s}$ and the $q_k$ are indexed out of the teacher-forced target pass that
already supplies the drafter's supervision. We define
\chg{\begin{equation}
\mathcal{L}_{\mathrm{dist}}
=
\frac{1}{|\mathcal{S}|}\sum_{s \in \mathcal{S}} \Omega_s ,
\qquad
\Omega_s
=
\sum_k
P_\theta\!\left(c_s=k \given c_{s-1}=g_{s-1}\right)
\relu\!\left(\log q_{g_s}-\log q_k\right),
\label{eq:distractor}
\end{equation}}
\chg{where $\Omega_s$ is the target-relative cost \method expects to pay at slot $s$ under its own
distribution, which are averaged over the supervised slots in $\mathcal{L}_{\mathrm{dist}}$.
Candidates the target ranks below $g_s$ are
penalized in proportion to the gap, and those it ranks above are not penalized at all.} When $g_s$
is the target's top candidate, this
term shares its gradient with a target cross-entropy over the candidate pool, and the rectifier
extends it safely to the rare positions where $g_s$ is not the target's top candidate, as
detailed in Appendix~\ref{app:derivations}.

We keep it alongside the
hard cross-entropy because the two supply complementary gradients. Cross-entropy makes \method
rank the ground truth first even when the target assigns it only modest probability, so it
remains the selected token however soft the target is. The distractor penalty complements this by pushing
the confusers the target disfavors further down. It is therefore cost-aware shaping of
cross-entropy, not a replacement, as detailed in Appendix~\ref{app:derivations}.

\subsection{Co-adapting the drafter and the reranker}
\label{sec:coadapt}

We tune \dflash jointly with \method, so that the drafter learns to expose candidate
pools and hidden representations from which \method can assemble longer accepted prefixes.
Joint training must cross the top-$\topk$ selection that forms the candidate pool, which is
not differentiable. We keep the selected candidate indices fixed and retain gradients through their
draft log-probabilities and the hidden states from which they were produced,\footnote{The
candidate token embeddings remain frozen.} both of which are inputs to \method. The
acceptance-oriented loss therefore reaches the \dflash backbone without differentiating pool
membership, and inference is unchanged.

The drafter keeps its own \dflash training loss, a per-slot block cross-entropy over the drafter
marginals that weights earlier slots more heavily \citep{dflash2025},
\begin{equation}
\mathcal{L}_{\text{\dflash}}
=
-\sum_{s=1}^{\block-1} w_s \, \log p_s(g_s),
\qquad
w_s = \exp\!\left(-\frac{s-1}{\beta}\right),
\label{eq:dflash}
\end{equation}
where $p_s$ is the drafter's marginal at slot $s$ and $\beta$ sets the decay. The drafter and
\method are trained jointly to minimize the drafter loss together with the two reranker terms,
\begin{equation}
\mathcal{L} \;=\; \mathcal{L}_{\text{\dflash}}
\;+\;
\gamma\left[
\mathcal{L}_{\mathrm{CE}}
\;+\;
\lambda\,\mathcal{L}_{\mathrm{dist}}
\right].
\label{eq:loss}
\end{equation}
Here $\gamma$ sets the weight of the reranker objective as a whole and $\lambda$ scales the
distractor penalty within it. These weights and the decay $\beta$ are given in
Appendix~\ref{app:impl}. \chg{Appendix~\ref{app:ablation} ablates the distractor penalty at a fixed
objective weight: it raises acceptance length on all nine benchmarks, and by $1.5\%$ on the
average.}

\subsection{Single-pass decoding}
\label{sec:decode}

Decoding has a parallel- and a sequential stage, and the network runs only
once. In the parallel stage, a single \method pass produces every outgoing and incoming
vector, including the anchor out vector. The block has $\block-1$ adjacent slot pairs, and within
each pair every candidate is matched against every candidate in the next slot. This gives
$\topk^2$ comparisons for each of the $\block\!-\!2$ interior pairs and $\topk$ for the anchor pair,
$(\block\!-\!2)\topk^2 + \topk$ matching terms in total. Evaluating
them reduces to one matrix multiplication per slot pair, batched across the full block, which
keeps the computation static in shape and well suited to kernel fusion and graph capture. In the sequential stage, the block is committed greedily from left to
right: at slot $s$ we select
\begin{equation}
\hat{c}_s \;=\; \argmax_k \, z_s[k]
\;=\; \argmax_k \, \Epair(s\!-\!1,\, \hat{c}_{s-1},\, k),
\end{equation}
conditioning each slot on the candidate just committed, a walk traced in
Figure~\ref{fig:teaser}(c). This greedy rule is exactly
the inference behavior the local objective in Eq.~\eqref{eq:localar} was trained
to optimize. An alternative is to commit the path that maximizes the sum of the matching terms.
Appendix~\ref{app:decode} recovers that path exactly with a dynamic program and finds it decodes
worse than the greedy rule. Only the argmax walk is sequential, and it touches no network
weights; the expensive computation is the single parallel pass. This is the key
distinction from iterative-refinement drafters, which invoke a network once per
position: \method confines all learning to one pass and reduces the
per-position work to an argmax over precomputed matching scores.

\section{Experiments}
\label{sec:experiments}

\subsection{Experimental setup}
\label{sec:exp-setup}

\paragraph{Models and evaluation.} We use frozen Qwen3-8B and Qwen3-4B targets
\citep{qwen3} with their reasoning mode disabled. Evaluation spans eight public datasets across three domains: math (GSM8K,
MATH-500, AIME 2025), code (HumanEval, MBPP, LiveCodeBench), and chat (Alpaca, MT-Bench).
We additionally report on SPEED-Bench \citep{abramovich2026speed}, hereafter SPEED, whose
qualitative split holds 880 prompts across eleven categories. We evaluate two decoding regimes: a
deterministic one with a greedy target, and a sampling one at temperature $1$ over the untruncated
target distribution, verified by exact rejection sampling so that the target distribution is
preserved and averaged over three seeds. We measure the acceptance length $\tau$, computed
block-weighted as the total number of accepted tokens divided by the number of verification
blocks, together with output throughput and the resulting speedup over the target's own autoregressive
decoding on the same hardware. \chg{Three ablations are deferred to
Appendices~\ref{app:ablation},~\ref{app:decode} and~\ref{app:conv}: the distractor penalty, the
greedy decode rule, and \dflashtwo's convolution placed in the \dflash backbone that precedes
\method.}

\paragraph{Baselines.} We compare \method against \chg{three} concurrent works to ours, Domino~\cite{domino2026},
DSpark~\cite{dspark2026} \chg{and \dflashtwo~\cite{dflash2_2026}}, and against the \dflash drafter it builds on.
\chg{All three tune the \dflash
drafter as they train their own head, as \method does. \dflashtwo's convolution acts on the drafter
rather than on the correlation head, so the two compose, and we ablate it on top of \method in
Appendix~\ref{app:conv}.} At temperature $1$
\chg{the verifier's rejection-sampling step reproduces the target distribution however the block was
drafted, so a drafter is free either to commit each slot to its own argmax or to sample its
proposal. Domino and \dflash only take the argmax, whereas \method, DSpark and
\dflashtwo do both. Table~\ref{tab:main-merged} reports the sampled proposal where a system has
one and the argmax otherwise, and Table~\ref{tab:main-greedy} reports every system under the
argmax. All systems are trained
on the same corpus, for the same number of epochs and at the same global
batch size and sequence length, under an identical configuration across systems and target
sizes, so acceptance length and throughput are directly comparable.}

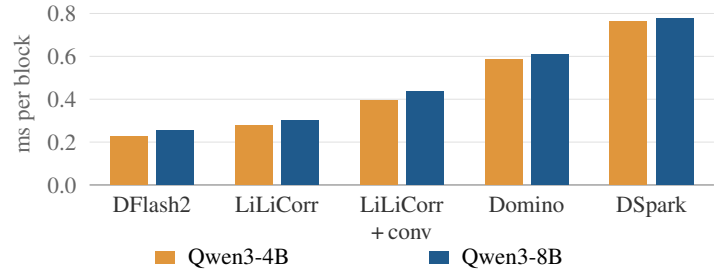
\begin{wrapfigure}[10]{r}{0.48\textwidth}
\centering
\begin{tikzpicture}[x=1mm,y=0.555mm,font=\tiny]
\definecolor{c8b}{RGB}{31,90,140}
\definecolor{c4b}{RGB}{224,150,60}
\draw[black!55,line width=.35pt] (0,0.00) -- (57.00,0.00);
\node[anchor=east,inner sep=1.5pt,text=black!60] at (-0.6,0.00) {0.0};
\draw[black!12,line width=.25pt] (0,7.06) -- (57.00,7.06);
\node[anchor=east,inner sep=1.5pt,text=black!60] at (-0.6,7.06) {0.2};
\draw[black!12,line width=.25pt] (0,14.12) -- (57.00,14.12);
\node[anchor=east,inner sep=1.5pt,text=black!60] at (-0.6,14.12) {0.4};
\draw[black!12,line width=.25pt] (0,21.18) -- (57.00,21.18);
\node[anchor=east,inner sep=1.5pt,text=black!60] at (-0.6,21.18) {0.6};
\draw[black!12,line width=.25pt] (0,28.24) -- (57.00,28.24);
\node[anchor=east,inner sep=1.5pt,text=black!60] at (-0.6,28.24) {0.8};
\fill[c4b] (1.88,0) rectangle (5.30,8.01);
\fill[c8b] (6.10,0) rectangle (9.52,9.04);
\node[anchor=north,align=center,inner sep=1.5pt,text=black!80] at (5.70,-0.8) {DFlash2};
\fill[c4b] (13.28,0) rectangle (16.70,9.74);
\fill[c8b] (17.50,0) rectangle (20.92,10.55);
\node[anchor=north,align=center,inner sep=1.5pt,text=black!80] at (17.10,-0.8) {LiLiCorr};
\fill[c4b] (24.68,0) rectangle (28.10,13.98);
\fill[c8b] (28.90,0) rectangle (32.32,15.39);
\node[anchor=north,align=center,inner sep=1.5pt,text=black!80] at (28.50,-0.8) {LiLiCorr\\+\,conv};
\fill[c4b] (36.08,0) rectangle (39.50,20.68);
\fill[c8b] (40.30,0) rectangle (43.72,21.56);
\node[anchor=north,align=center,inner sep=1.5pt,text=black!80] at (39.90,-0.8) {Domino};
\fill[c4b] (47.48,0) rectangle (50.90,27.00);
\fill[c8b] (51.70,0) rectangle (55.12,27.49);
\node[anchor=north,align=center,inner sep=1.5pt,text=black!80] at (51.30,-0.8) {DSpark};
\node[anchor=south,rotate=90,inner sep=1pt,text=black!60] at (-4.6,15.00) {ms per block};
\fill[c4b] (6.0,-13.00) rectangle (7.8,-11.00);
\node[anchor=west,inner sep=1.5pt] at (8.4,-12.00) {Qwen3-4B};
\fill[c8b] (31.0,-13.00) rectangle (32.8,-11.00);
\node[anchor=west,inner sep=1.5pt] at (33.4,-12.00) {Qwen3-8B};
\end{tikzpicture}
\caption{\chg{Per-block latency of each approach.}}
\label{fig:head-cost}
\end{wrapfigure}

\paragraph{Implementation details.} We serve every system on the same deployment-representative
SGLang backend, with block size 16 on a single H100 and concurrency one
unless stated otherwise. \chg{Every speculative head is folded into the drafter and optimized with
fused kernels and graph capture. Fully
optimized, the \method head costs $0.299$ ms per
block, against Domino's $0.611$ and DSpark's $0.779$, and accounts for about $3\%$ of the per-block
wall on Qwen3-8B and $4.5\%$ on the cheaper Qwen3-4B block (Figure~\ref{fig:head-cost}).
\dflashtwo's head and convolution together are slightly cheaper, at $0.256$ ms.} \chg{Table~\ref{tab:cost}
presents these as shares of the block, and} the per-stage
decomposition behind them is in Appendix~\ref{app:impl}.

\chg{Acceptance length is reproducible to the digits we print, but throughput is not: repeated runs
of one system on one node at concurrency one differ by up to $0.94\%$. We therefore set the noise
floor at $1\%$, reading a throughput difference of at most that as a tie and counting a win only
where a system leads every other by more.}

\chg{\paragraph{Training data.} All systems are trained on roughly $1.4$M examples
drawn from the code, mathematics, STEM, and general-chat splits of the Nemotron Post-Training
Dataset V2 \citep{nemotronpost}, with the multilingual splits excluded and every assistant turn
regenerated with the target model (Appendix~\ref{app:impl}).}

\subsection{Main results}
\label{sec:exp-main}

Table~\ref{tab:main-merged} reports speedup over autoregressive decoding on the
nine benchmarks, for both target sizes and both decoding regimes; the acceptance lengths and
absolute throughputs behind it are deferred to Appendix~\ref{app:full-results}. \chg{The four
draft-time methods reach comparable acceptance lengths, so throughput separates them. Under greedy decoding \method is fastest in twelve of the eighteen settings, ties in four and
trails in two, both deficits being HumanEval against \dflashtwo. Against the \dflash drafter, which
reranks nothing, \method improves on both axes everywhere, raising acceptance length by $13$ to
$19\%$ and throughput by $6$ to $14\%$. Tables~\ref{tab:speed-8b} and~\ref{tab:speed-4b} resolve
SPEED into its eleven categories, in all forty-four of which \method serves faster than every
system compared, and by its widest acceptance margin on multilingual. It is the only domain held
out of training, which suggests a capacity to generalize out of distribution.}

\chg{Sampling does not reorder the systems. At temperature one accepted prefixes shorten for every
system, as expected once the target no longer commits to its own argmax. Among argmax proposals \method is fastest at all eighteen settings, and among
sampled ones at fifteen, tying at two and trailing at one. DSpark accepts longer prefixes than \method
at 8B, by $3.1\%$ on the sampled average, and still serves slower: our cheaper head completes
$7.2\%$ more verification blocks per second, which more than offsets the shorter prefix
(Table~\ref{tab:main-8b}).}

\begin{table}[!t]
\centering
\footnotesize
\setlength{\tabcolsep}{1.6pt}
\caption{\chg{Speedup over autoregressive decoding, single H100, concurrency one. $T\!=\!1$ entries are
means over three seeds and the Average is unweighted. Fastest per row and target size in bold, any
system within the $1\%$ tie band underlined.}}
\label{tab:main-merged}
\begin{tabular*}{\textwidth}{@{}c@{\hspace{3pt}}l@{\extracolsep{\fill}}|cc|cc|cc|cc|cc|cc@{}}
\toprule
& & \multicolumn{2}{c|}{AR (TPS)} & \multicolumn{2}{c|}{\method} & \multicolumn{2}{c|}{\dflashtwo} & \multicolumn{2}{c|}{Domino} & \multicolumn{2}{c|}{DSpark} & \multicolumn{2}{c}{\dflash} \\
& Benchmark & 8B & 4B & 8B & 4B & 8B & 4B & 8B & 4B & 8B & 4B & 8B & 4B \\
\midrule
\multirow{10}{*}{\rotatebox[origin=c]{90}{\scriptsize \textit{Greedy}, $T\!=\!0$}}
& GSM8K & 149 & 234 & \textbf{5.32}$\times$ & \textbf{5.00}$\times$ & 5.15$\times$ & 4.90$\times$ & 4.94$\times$ & 4.56$\times$ & 4.95$\times$ & 4.50$\times$ & 4.66$\times$ & 4.43$\times$ \\
& MATH-500 & 150 & 235 & \textbf{6.57}$\times$ & \underline{6.13}$\times$ & \underline{6.54}$\times$ & \textbf{6.14}$\times$ & 6.29$\times$ & 5.78$\times$ & 6.20$\times$ & 5.55$\times$ & 5.93$\times$ & 5.60$\times$ \\
& AIME 2025 & 149 & 233 & \textbf{6.02}$\times$ & \underline{5.89}$\times$ & 5.90$\times$ & \textbf{5.90}$\times$ & 5.75$\times$ & 5.53$\times$ & 5.64$\times$ & 5.30$\times$ & 5.42$\times$ & 5.37$\times$ \\
& HumanEval & 146 & 226 & 4.02$\times$ & 3.90$\times$ & \textbf{4.11}$\times$ & \textbf{3.99}$\times$ & 3.89$\times$ & 3.73$\times$ & 3.99$\times$ & 3.67$\times$ & 3.79$\times$ & 3.64$\times$ \\
& MBPP & 150 & 235 & \textbf{4.16}$\times$ & \textbf{3.91}$\times$ & 4.08$\times$ & 3.83$\times$ & 3.93$\times$ & 3.63$\times$ & 3.99$\times$ & 3.58$\times$ & 3.72$\times$ & 3.50$\times$ \\
& LiveCodeBench & 148 & 218 & \textbf{5.43}$\times$ & \underline{5.53}$\times$ & 5.35$\times$ & \textbf{5.58}$\times$ & 5.12$\times$ & 5.18$\times$ & 5.18$\times$ & 5.16$\times$ & 4.94$\times$ & 5.03$\times$ \\
& AlpacaEval & 150 & 236 & \textbf{2.70}$\times$ & \textbf{2.58}$\times$ & 2.58$\times$ & 2.48$\times$ & 2.59$\times$ & 2.44$\times$ & 2.52$\times$ & 2.35$\times$ & 2.46$\times$ & 2.39$\times$ \\
& MT-Bench & 150 & 234 & \textbf{2.90}$\times$ & \textbf{2.80}$\times$ & 2.77$\times$ & 2.69$\times$ & 2.81$\times$ & 2.62$\times$ & 2.76$\times$ & 2.55$\times$ & 2.64$\times$ & 2.58$\times$ \\
& SPEED & 149 & 232 & \textbf{3.61}$\times$ & \textbf{3.50}$\times$ & 3.44$\times$ & 3.40$\times$ & 3.38$\times$ & 3.25$\times$ & 3.37$\times$ & 3.16$\times$ & 3.24$\times$ & 3.22$\times$ \\
\cmidrule(lr){2-14}
& Average & 149 & 232 & \textbf{4.53}$\times$ & \textbf{4.36}$\times$ & 4.44$\times$ & \underline{4.32}$\times$ & 4.30$\times$ & 4.08$\times$ & 4.29$\times$ & 3.98$\times$ & 4.09$\times$ & 3.97$\times$ \\
\addlinespace[4pt]
\multirow{10}{*}{\rotatebox[origin=c]{90}{\scriptsize \textit{Sampling}, $T\!=\!1$}}
& GSM8K & 148 & 230 & \textbf{4.87}$\times$ & \textbf{4.62}$\times$ & 3.88$\times$ & 3.45$\times$ & 4.31$\times$ & 4.10$\times$ & 4.53$\times$ & 4.09$\times$ & 4.13$\times$ & 4.01$\times$ \\
& MATH-500 & 148 & 228 & \underline{5.01}$\times$ & \underline{4.48}$\times$ & 4.20$\times$ & 3.72$\times$ & 4.40$\times$ & 4.17$\times$ & \textbf{5.03}$\times$ & \textbf{4.52}$\times$ & 4.30$\times$ & 4.07$\times$ \\
& AIME 2025 & 147 & 228 & 3.99$\times$ & \textbf{3.93}$\times$ & 3.38$\times$ & 2.99$\times$ & 3.43$\times$ & 3.29$\times$ & \textbf{4.16}$\times$ & 3.77$\times$ & 3.47$\times$ & 3.28$\times$ \\
& HumanEval & 143 & 221 & \textbf{4.51}$\times$ & \textbf{4.12}$\times$ & 3.77$\times$ & 3.32$\times$ & 4.24$\times$ & 3.89$\times$ & 4.09$\times$ & 3.80$\times$ & 4.13$\times$ & 3.97$\times$ \\
& MBPP & 148 & 231 & \textbf{3.73}$\times$ & \textbf{3.68}$\times$ & 3.06$\times$ & 2.74$\times$ & 3.34$\times$ & 3.29$\times$ & 3.64$\times$ & 3.27$\times$ & 3.26$\times$ & 3.24$\times$ \\
& LiveCodeBench & 146 & 226 & \textbf{4.28}$\times$ & \textbf{3.91}$\times$ & 3.29$\times$ & 2.88$\times$ & 3.88$\times$ & 3.56$\times$ & 4.05$\times$ & 3.50$\times$ & 3.78$\times$ & 3.51$\times$ \\
& AlpacaEval & 148 & 229 & \textbf{2.36}$\times$ & \textbf{2.30}$\times$ & 1.90$\times$ & 1.64$\times$ & 2.08$\times$ & 2.13$\times$ & 2.24$\times$ & 2.06$\times$ & 2.05$\times$ & 2.11$\times$ \\
& MT-Bench & 148 & 230 & \textbf{2.67}$\times$ & \textbf{2.59}$\times$ & 2.13$\times$ & 1.85$\times$ & 2.39$\times$ & 2.34$\times$ & 2.51$\times$ & 2.31$\times$ & 2.29$\times$ & 2.33$\times$ \\
& SPEED & 147 & 228 & \textbf{2.77}$\times$ & \textbf{2.81}$\times$ & 2.23$\times$ & 2.06$\times$ & 2.41$\times$ & 2.41$\times$ & 2.63$\times$ & 2.50$\times$ & 2.36$\times$ & 2.40$\times$ \\
\cmidrule(lr){2-14}
& Average & 147 & 228 & \textbf{3.80}$\times$ & \textbf{3.60}$\times$ & 3.10$\times$ & 2.74$\times$ & 3.39$\times$ & 3.24$\times$ & 3.65$\times$ & 3.31$\times$ & 3.31$\times$ & 3.21$\times$ \\
\bottomrule
\end{tabular*}
\end{table}

\subsection{Scaling with concurrency}
\label{sec:exp-concurrency}

Deployed systems run batched, a regime in which decoding moves from memory-bandwidth bound
towards compute bound and the cost of speculation is weighed differently. SPEED provides a second,
throughput split for this setting, which holds the input length fixed, generates \chg{$1{,}024$
tokens per request} and varies serving concurrency
\citep{abramovich2026speed}. We evaluate on its $1$K and $2$K input-length blocks, in each of their
three output-entropy tiers. \chg{The tiers group prompts by how predictable their continuations
are: coding and mathematics sit in the low-entropy tier and are easier to draft, roleplay and
writing in the high-entropy tier and are harder.} Every drafter is trained on sequences of \chg{$3{,}120$} tokens
(Table~\ref{tab:hparams}), and these are the two blocks whose input and generation together stay
within that length.

Table~\ref{tab:concurrency} covers concurrencies from $c\!=\!1$ to $c\!=\!32$. \chg{\method attains
both the highest throughput and the longest accepted prefix at every one of the thirty-six
operating points, none of them inside the tie band, leading on acceptance by $1.1$ to $5.7\%$ over
the strongest baseline in each cell. Its accepted prefix barely moves with the batch, by between
$-1.4\%$ and $+0.5\%$ from $c\!=\!1$ to $c\!=\!32$.} Speedups compress for all systems as the batch
grows, from \chg{$2.1$--$3.8\times$} at $c\!=\!1$ to \chg{$1.3$--$1.8\times$} at $c\!=\!32$, and
the ordering is preserved throughout. Acceptance
lengths and absolute throughputs are given in Appendix~\ref{app:isl}.

\begin{table}[!t]
\centering
\footnotesize
\setlength{\tabcolsep}{1.2pt}
\caption{\chg{Speedup over autoregressive decoding as serving concurrency $c$ grows, on the $1$K and $2$K
blocks of the SPEED throughput split \citep{abramovich2026speed}, Qwen3-8B, greedy. Tiers are
low (L), mixed (M) and high (H) output entropy. Fastest per cell in bold, and no cell falls inside
the tie band. Acceptance lengths are in Table~\ref{tab:isl-short}.}}
\label{tab:concurrency}
\begin{tabular*}{\textwidth}{@{}c@{\hspace{2pt}}c@{\extracolsep{\fill}}|ccc|ccc|ccc|ccc|ccc@{}}
\toprule
& & \multicolumn{3}{c|}{\method} & \multicolumn{3}{c|}{\dflashtwo} & \multicolumn{3}{c|}{Domino} & \multicolumn{3}{c|}{DSpark} & \multicolumn{3}{c}{\dflash} \\
& $c$ & L & M & H & L & M & H & L & M & H & L & M & H & L & M & H \\
\midrule
\multirow{6}{*}{\rotatebox[origin=c]{90}{\scriptsize \textit{$1K$}}}
& 1 & \textbf{3.48}$\times$ & \textbf{3.55}$\times$ & \textbf{2.31}$\times$ & 3.26$\times$ & 3.39$\times$ & 2.16$\times$ & 3.13$\times$ & 3.32$\times$ & 2.17$\times$ & 3.19$\times$ & 3.31$\times$ & 2.13$\times$ & 3.11$\times$ & 3.18$\times$ & 2.13$\times$ \\
& 2 & \textbf{3.37}$\times$ & \textbf{3.51}$\times$ & \textbf{2.33}$\times$ & 3.17$\times$ & 3.36$\times$ & 2.16$\times$ & 3.05$\times$ & 3.28$\times$ & 2.20$\times$ & 3.11$\times$ & 3.25$\times$ & 2.12$\times$ & 3.05$\times$ & 3.16$\times$ & 2.12$\times$ \\
& 4 & \textbf{3.18}$\times$ & \textbf{3.31}$\times$ & \textbf{2.24}$\times$ & 3.03$\times$ & 3.21$\times$ & 2.13$\times$ & 2.88$\times$ & 3.14$\times$ & 2.11$\times$ & 2.94$\times$ & 3.11$\times$ & 2.06$\times$ & 2.88$\times$ & 3.00$\times$ & 2.08$\times$ \\
& 8 & \textbf{2.79}$\times$ & \textbf{2.95}$\times$ & \textbf{2.08}$\times$ & 2.65$\times$ & 2.81$\times$ & 1.93$\times$ & 2.52$\times$ & 2.80$\times$ & 1.93$\times$ & 2.59$\times$ & 2.75$\times$ & 1.91$\times$ & 2.54$\times$ & 2.69$\times$ & 1.93$\times$ \\
& 16 & \textbf{2.32}$\times$ & \textbf{2.44}$\times$ & \textbf{1.81}$\times$ & 2.19$\times$ & 2.39$\times$ & 1.69$\times$ & 2.16$\times$ & 2.36$\times$ & 1.73$\times$ & 2.19$\times$ & 2.33$\times$ & 1.67$\times$ & 2.15$\times$ & 2.28$\times$ & 1.67$\times$ \\
& 32 & \textbf{1.68}$\times$ & \textbf{1.81}$\times$ & \textbf{1.33}$\times$ & 1.61$\times$ & 1.76$\times$ & 1.27$\times$ & 1.60$\times$ & 1.77$\times$ & 1.30$\times$ & 1.62$\times$ & 1.76$\times$ & 1.26$\times$ & 1.56$\times$ & 1.69$\times$ & 1.25$\times$ \\
\addlinespace[4pt]
\multirow{6}{*}{\rotatebox[origin=c]{90}{\scriptsize \textit{$2K$}}}
& 1 & \textbf{3.83}$\times$ & \textbf{3.68}$\times$ & \textbf{2.31}$\times$ & 3.54$\times$ & 3.52$\times$ & 2.17$\times$ & 3.39$\times$ & 3.41$\times$ & 2.17$\times$ & 3.48$\times$ & 3.41$\times$ & 2.13$\times$ & 3.37$\times$ & 3.30$\times$ & 2.13$\times$ \\
& 2 & \textbf{3.72}$\times$ & \textbf{3.60}$\times$ & \textbf{2.33}$\times$ & 3.46$\times$ & 3.42$\times$ & 2.19$\times$ & 3.32$\times$ & 3.35$\times$ & 2.19$\times$ & 3.37$\times$ & 3.32$\times$ & 2.13$\times$ & 3.29$\times$ & 3.23$\times$ & 2.17$\times$ \\
& 4 & \textbf{3.40}$\times$ & \textbf{3.32}$\times$ & \textbf{2.24}$\times$ & 3.20$\times$ & 3.20$\times$ & 2.12$\times$ & 3.09$\times$ & 3.10$\times$ & 2.11$\times$ & 3.15$\times$ & 3.10$\times$ & 2.07$\times$ & 3.06$\times$ & 3.04$\times$ & 2.10$\times$ \\
& 8 & \textbf{2.97}$\times$ & \textbf{2.90}$\times$ & \textbf{2.04}$\times$ & 2.77$\times$ & 2.78$\times$ & 1.93$\times$ & 2.67$\times$ & 2.75$\times$ & 1.93$\times$ & 2.74$\times$ & 2.69$\times$ & 1.88$\times$ & 2.66$\times$ & 2.65$\times$ & 1.91$\times$ \\
& 16 & \textbf{2.39}$\times$ & \textbf{2.39}$\times$ & \textbf{1.80}$\times$ & 2.32$\times$ & 2.32$\times$ & 1.71$\times$ & 2.24$\times$ & 2.31$\times$ & 1.73$\times$ & 2.27$\times$ & 2.25$\times$ & 1.69$\times$ & 2.22$\times$ & 2.21$\times$ & 1.67$\times$ \\
& 32 & \textbf{1.83}$\times$ & \textbf{1.82}$\times$ & \textbf{1.39}$\times$ & 1.75$\times$ & 1.78$\times$ & 1.33$\times$ & 1.73$\times$ & 1.79$\times$ & 1.36$\times$ & 1.75$\times$ & 1.77$\times$ & 1.34$\times$ & 1.70$\times$ & 1.71$\times$ & 1.30$\times$ \\
\bottomrule
\end{tabular*}
\end{table}

%

\section{Conclusion}
\method correlates the per-slot marginals of a parallel block drafter in a single network
pass. Assigning each candidate an \emph{in} and an \emph{out} vector reduces the correlation to
batched matrix products, leaving only a greedy walk that touches no network weights. Trained
jointly with the drafter, \method improves on the vanilla \dflash drafter on every benchmark we
test, and a draft of comparable quality at a fraction of the cost of correcting one slot at a time
makes it the faster system almost everywhere we measure.

One limitation is inherent to the design: \method reranks only the top-$\topk$ tokens the
drafter proposes, so a slot whose correct token is absent from that pool cannot be recovered, and
the achievable acceptance length is bounded by the drafter's coverage. \chg{A second follows from scoring the block
through adjacent pairs alone. Each candidate is represented from the whole block, so evidence from
distant slots reaches a pair score, but only after being compressed into that one number. Coupling
distant slots explicitly is a natural next step, and whether it repays its cost remains open.}

\bibliographystyle{iclr2027_conference}
\bibliography{references}

\begin{thebibliography}{22}
\providecommand{\natexlab}[1]{#1}
\providecommand{\url}[1]{\texttt{#1}}
\expandafter\ifx\csname urlstyle\endcsname\relax
  \providecommand{\doi}[1]{doi: #1}\else
  \providecommand{\doi}{doi: \begingroup \urlstyle{rm}\Url}\fi

\bibitem[Abramovich et~al.(2026)Abramovich, Ashkenazi, Putterman, Chislett,
  Mitra, Rouhani, Zilberstein, and Geifman]{abramovich2026speed}
Talor Abramovich, Maor Ashkenazi, Izzy Putterman, Benjamin Chislett, Tiyasa
  Mitra, Bita~Darvish Rouhani, Ran Zilberstein, and Yonatan Geifman.
\newblock {SPEED}-bench: A unified and diverse benchmark for speculative
  decoding.
\newblock In \emph{Forty-third International Conference on Machine Learning},
  2026.
\newblock URL \url{https://openreview.net/forum?id=Rl2uQlCoQX}.

\bibitem[Cai et~al.(2024)Cai, Li, Geng, Peng, Lee, Chen, and
  Dao]{cai2024medusa}
Tianle Cai, Yuhong Li, Zhengyang Geng, Hongwu Peng, Jason~D. Lee, Deming Chen,
  and Tri Dao.
\newblock Medusa: Simple {LLM} inference acceleration framework with multiple
  decoding heads.
\newblock In \emph{Forty-first International Conference on Machine Learning},
  2024.
\newblock URL \url{https://openreview.net/forum?id=PEpbUobfJv}.

\bibitem[Chen et~al.(2023)Chen, Borgeaud, Irving, Lespiau, Sifre, and
  Jumper]{chen2023accelerating}
Charlie Chen, Sebastian Borgeaud, Geoffrey Irving, Jean-Baptiste Lespiau,
  Laurent Sifre, and John Jumper.
\newblock Accelerating large language model decoding with speculative sampling.
\newblock \emph{arXiv preprint arXiv:2302.01318}, 2023.

\bibitem[Chen et~al.(2026)Chen, Liang, and Liu]{dflash2025}
Jian Chen, Yesheng Liang, and Zhijian Liu.
\newblock {DF}lash: Block diffusion for flash speculative decoding.
\newblock In \emph{Forty-third International Conference on Machine Learning},
  2026.
\newblock URL \url{https://openreview.net/forum?id=Oz335dV48X}.

\bibitem[Cheng et~al.(2026)Cheng, Yu, Shao, Li, Xiong, Qian, Zhu, Ma, Zhang,
  Ye, et~al.]{dspark2026}
Xin Cheng, Xingkai Yu, Chenze Shao, Jiashi Li, Yunfan Xiong, Yi~Qian, Jiaqi
  Zhu, Shirong Ma, Xiaokang Zhang, Jiasheng Ye, et~al.
\newblock Dspark: Confidence-scheduled speculative decoding with
  semi-autoregressive generation.
\newblock \emph{arXiv preprint arXiv:2607.05147}, 2026.

\bibitem[Fu et~al.(2026)Fu, Whalen, Garg, Wu, Khadkevich, Oswald, Xie, Egert,
  Sreenivas, Diao, et~al.]{nld2026}
Yonggan Fu, Lexington Whalen, Abhinav Garg, Chengyue Wu, Maksim Khadkevich,
  Nicolai Oswald, Enze Xie, Daniel Egert, Sharath~Turuvekere Sreenivas, Shizhe
  Diao, et~al.
\newblock Nemotron-labs-diffusion: A tri-mode language model unifying
  autoregressive, diffusion, and self-speculation decoding.
\newblock \emph{arXiv preprint arXiv:2607.05722}, 2026.

\bibitem[Huang et~al.(2026)Huang, Zhang, Zhang, Lin, Xu, and Zhang]{domino2026}
Jianuo Huang, Yaojie Zhang, Qituan Zhang, Hao Lin, Hanlin Xu, and Linfeng
  Zhang.
\newblock Domino: Decoupling causal modeling from autoregressive drafting in
  speculative decoding.
\newblock \emph{arXiv preprint arXiv:2605.29707}, 2026.

\bibitem[{Inco AI}(2026)]{dflash2_2026}
{Inco AI}.
\newblock {DFlash 2}.
\newblock Technical blog, August 2026.
\newblock URL \url{https://inco.ai/blog/dflash2/}.

\bibitem[Leviathan et~al.(2023)Leviathan, Kalman, and
  Matias]{leviathan2023speculative}
Yaniv Leviathan, Matan Kalman, and Yossi Matias.
\newblock Fast inference from transformers via speculative decoding.
\newblock In \emph{International conference on machine learning}, pages
  19274--19286. PMLR, 2023.

\bibitem[Li et~al.(2024{\natexlab{a}})Li, Wei, Zhang, and Zhang]{li2024eagle}
Yuhui Li, Fangyun Wei, Chao Zhang, and Hongyang Zhang.
\newblock {EAGLE}: Speculative sampling requires rethinking feature
  uncertainty.
\newblock In \emph{Forty-first International Conference on Machine Learning},
  2024{\natexlab{a}}.
\newblock URL \url{https://openreview.net/forum?id=1NdN7eXyb4}.

\bibitem[Li et~al.(2024{\natexlab{b}})Li, Wei, Zhang, and Zhang]{li2024eagle2}
Yuhui Li, Fangyun Wei, Chao Zhang, and Hongyang Zhang.
\newblock Eagle-2: Faster inference of language models with dynamic draft
  trees.
\newblock In \emph{Proceedings of the 2024 conference on empirical methods in
  natural language processing}, pages 7421--7432, 2024{\natexlab{b}}.

\bibitem[Li et~al.(2026)Li, Wei, Zhang, and Zhang]{li2025eagle3}
Yuhui Li, Fangyun Wei, Chao Zhang, and Hongyang Zhang.
\newblock Eagle-3: Scaling up inference acceleration of large language models
  via training-time test.
\newblock \emph{Advances in Neural Information Processing Systems},
  38:\penalty0 136737--136756, 2026.

\bibitem[Nathawani et~al.(2025)Nathawani, Ding, Lavrukhin, Gitman, Majumdar,
  Bakhturina, Ginsburg, and Polak~Scowcroft]{nemotronpost}
Dhruv Nathawani, Shuoyang Ding, Vitaly Lavrukhin, Igor Gitman, Somshubra
  Majumdar, Evelina Bakhturina, Boris Ginsburg, and Jane Polak~Scowcroft.
\newblock {Nemotron-Post-Training-Dataset-v2}.
\newblock Version 2.0, NVIDIA, August 2025.
\newblock URL
  \url{https://huggingface.co/datasets/nvidia/Nemotron-Post-Training-Dataset-v2}.

\bibitem[Nie et~al.(2026)Nie, Zhu, You, Zhang, Ou, Hu, Zhou, Lin, Wen, and
  Li]{nie2025llada}
Shen Nie, Fengqi Zhu, Zebin You, Xiaolu Zhang, Jingyang Ou, Jun Hu, Jun Zhou,
  Yankai Lin, Ji-Rong Wen, and Chongxuan Li.
\newblock Large language diffusion models.
\newblock \emph{Advances in Neural Information Processing Systems},
  38:\penalty0 50608--50646, 2026.

\bibitem[Oda et~al.(2026)Oda, Mathieu, Knyazhitskiy, and Chakhvadze]{tfm2026}
Yuma Oda, Ryan Mathieu, Roman Knyazhitskiy, and Artur Chakhvadze.
\newblock Trees from marginals: Autoregressive drafting with factorized priors.
\newblock \emph{arXiv preprint arXiv:2607.06763}, 2026.

\bibitem[Ringel and Romano(2026)]{ddtree}
Liran Ringel and Yaniv Romano.
\newblock Accelerating speculative decoding with block diffusion draft trees.
\newblock \emph{arXiv preprint arXiv:2604.12989}, 2026.

\bibitem[Sahoo et~al.(2024)Sahoo, Arriola, Schiff, Gokaslan, Marroquin, Chiu,
  Rush, and Kuleshov]{sahoo2024mdlm}
Subham~S Sahoo, Marianne Arriola, Yair Schiff, Aaron Gokaslan, Edgar Marroquin,
  Justin~T Chiu, Alexander Rush, and Volodymyr Kuleshov.
\newblock Simple and effective masked diffusion language models.
\newblock \emph{Advances in Neural Information Processing Systems},
  37:\penalty0 130136--130184, 2024.

\bibitem[Sun et~al.(2019)Sun, Li, Wang, He, Lin, and Deng]{sun2019fast}
Zhiqing Sun, Zhuohan Li, Haoqing Wang, Di~He, Zi~Lin, and Zhihong Deng.
\newblock Fast structured decoding for sequence models.
\newblock \emph{Advances in Neural Information Processing Systems}, 32, 2019.

\bibitem[Viterbi(1967)]{viterbi1967}
Andrew Viterbi.
\newblock Error bounds for convolutional codes and an asymptotically optimum
  decoding algorithm.
\newblock \emph{IEEE transactions on Information Theory}, 13\penalty0
  (2):\penalty0 260--269, 1967.

\bibitem[Wu et~al.(2026)Wu, Yao, Qi, Zheng, Wang, Ma, Liao, Lakkaraju, Li, and
  Du]{dpace2026}
Tianyu Wu, Yu~Yao, Zhenting Qi, Han Zheng, Zhuohan Wang, Haoran Ma, Lawrence
  Liao, Himabindu Lakkaraju, Ju~Li, and Yilun Du.
\newblock D-pace: Dynamic position-aware cross-entropy for parallel speculative
  drafting.
\newblock \emph{arXiv preprint arXiv:2605.18810}, 2026.

\bibitem[Yang et~al.(2025)Yang, Li, Yang, Zhang, Hui, Zheng, Yu, Gao, Huang,
  Lv, et~al.]{qwen3}
An~Yang, Anfeng Li, Baosong Yang, Beichen Zhang, Binyuan Hui, Bo~Zheng, Bowen
  Yu, Chang Gao, Chengen Huang, Chenxu Lv, et~al.
\newblock Qwen3 technical report.
\newblock \emph{arXiv preprint arXiv:2505.09388}, 2025.

\bibitem[Ye et~al.(2025)Ye, Xie, Zheng, Gao, Wu, Jiang, Li, and
  Kong]{dream2025}
Jiacheng Ye, Zhihui Xie, Lin Zheng, Jiahui Gao, Zirui Wu, Xin Jiang, Zhenguo
  Li, and Lingpeng Kong.
\newblock Dream 7b: Diffusion large language models.
\newblock \emph{arXiv preprint arXiv:2508.15487}, 2025.

\end{thebibliography}

\appendix
\section{Implementation and Training Details}
\label{app:impl}

\paragraph{Architecture.}
Each candidate enters the head as one vector of the head's hidden width, formed by summing
four terms. The first is the target's input embedding of the candidate token, held frozen and
projected down. The second is the drafter's hidden state at that slot, projected down and shared by
the slot's $\topk$ candidates. The third is a two-layer network over five scalars describing the
drafter's confidence in the candidate, all read from its distribution at that slot: the
log-probability and the probability under the full vocabulary, the log-probability gap to the
slot's most likely token, the rank normalized to $[0,1]$, and an indicator for being the slot's top
choice. The fourth is a pair of learned tables, one
indexed by slot and one by rank. The target state that every candidate is later fused with is
formed separately: the target's hidden states that produced the last verified token, collapsed
across five captured layers, normalized, and projected to the same width.

The head itself is a two-layer bidirectional Transformer over the
$(\block\!-\!1)\times\topk$ candidate lattice, attending over all of it at once, with hidden width
$1024$, eight attention heads, and a feed-forward width of $2048$, using RMSNorm and SiLU
activations. Beyond the slot table, position enters through two attention biases learned per head
and added to the attention logits, one indexed by the signed slot distance between a pair and one
marking pairs of candidates that occupy the same slot. Only after the Transformer is each candidate
combined with that target state, by concatenating the candidate state, the target state, and their
elementwise product, and projecting the result back down to the head's width through a single
linear layer followed by a SiLU. This is the one concatenation in the head, everything else being a
sum. Two linear heads then emit the out and in vectors of
Section~\ref{sec:lattice}, each of dimension $d_e$ and normalized to unit norm, and a third emits
the anchor's single out vector. Table~\ref{tab:hparams} presents the settings.

\chg{\paragraph{Candidate pool size.} We fixed $\topk$ at $8$ after trying smaller and larger
pools. At $\topk\!=\!4$ fewer slots contain the ground-truth token, so the reranker has less to
recover; at $\topk\!=\!16$ coverage improves, but the added candidates are ones the drafter already
ranks low and the wider lattice costs more per block. Both served slower than $\topk\!=\!8$ in our
runs.}

\chg{\paragraph{Training corpus.} For each conversation we keep the original system and user
messages and regenerate every assistant turn with the target model, decoding greedily with
reasoning disabled and a cap of $4{,}096$ completion tokens.}

\paragraph{Objective.}
The \dflash cross-entropy of Eq.~\eqref{eq:dflash} is applied at every slot, whereas the
\method terms are evaluated only on the oracle prefix, the maximal run of leading slots in which
the ground-truth token is present in the candidate pool. Slots beyond the first coverage miss
contribute to none of the three terms.

\paragraph{Serving.}
We fuse top-$\topk$ candidate extraction and vocabulary-wide logit normalization into a single
kernel, compile the matching and selection step, and use wide CUDA graphs that fold candidate
pooling and reranking into a single replay. The same level of optimization is applied to Domino's
correction loop. \chg{Table~\ref{tab:cost} gives the resulting per-block cost of each head and is
the source of the figures quoted in Section~\ref{sec:exp-setup}.}

\paragraph{\chg{Isolating the head.}}
\chg{Table~\ref{tab:cost} is obtained by subtraction. Every benchmark run under greedy decoding
reports throughput and
acceptance length, and their ratio is the wall-clock time of one verification block. The \dflash
control is the same system with no reranking at all, and its block takes $8.87$ ms on Qwen3-8B and
$5.88$ ms on Qwen3-4B, so subtracting it from another system's block time leaves what that system's
head costs. All six run together in each allocation, so the choice of machine cancels in the
subtraction. We take the median block time over eight of the nine benchmarks, excluding HumanEval,
whose short outputs let prefill dominate, and subtract one median from the other. Two things follow
from measuring a difference rather than a stage. The figure is net of the per-slot argmax the
control performs instead, which is the cost a deployment actually pays for reranking. And it
resolves about $0.03$ ms, the spread across three independent reads of the convolution's per-block
cost, with per-benchmark head costs varying by up to $\pm0.1$ ms around the median.}

\begin{table}[t]
\centering
\small
\caption{\chg{Per-block cost of each speculative head, obtained by the subtraction described
above. For \dflashtwo and \methodc the figure covers the drafter-side convolution as well as the
head.}}
\label{tab:cost}
\chg{\begin{tabular}{lcccc}
\toprule
 & \multicolumn{2}{c}{Qwen3-8B} & \multicolumn{2}{c}{Qwen3-4B} \\
\cmidrule(lr){2-3}\cmidrule(lr){4-5}
System & Head (ms) & Share of block & Head (ms) & Share of block \\
\midrule
\dflashtwo & 0.256 & 2.8\% & 0.227 & 3.7\% \\
\method    & 0.299 & 3.3\% & 0.276 & 4.5\% \\
\methodc   & 0.436 & 4.7\% & 0.396 & 6.3\% \\
Domino     & 0.611 & 6.5\% & 0.586 & 9.1\% \\
DSpark     & 0.779 & 8.1\% & 0.765 & 11.5\% \\
\bottomrule
\end{tabular}}
\end{table}

\chg{\paragraph{Precision.}
The drafter is trained with a full-precision master copy of its weights, and the optimizer moments
are kept in full precision throughout. Two details follow from how that copy is handled. Some
training runs stopped and resumed. On resume the drafter's weights are reloaded at the target
model's precision, so the master copy is round-tripped through bf16 at that point, in resumed and
uninterrupted runs alike. Where a resumed run re-covered steps it had already taken, the recomputed
loss values matched the earlier pass to the precision logged. Separately, the drafter is
initialized in full precision, but in the runs reported here that initialization was rounded to the
target's precision before training began, so that arms differing only in precision started from
numerically identical weights. The rounding moves the initialization only below bf16 resolution and
leaves first-step loss and gradients unchanged.}

\paragraph{Baselines.}
Domino is served in its widened configuration, \chg{trained at $512$ anchors rather than its
released $256$.} Its correction head is a recurrent cell over the drafter's
hidden width followed by a vocabulary projection. DSpark is served
in a width-matched configuration, and at temperature $1$ under each of
\chg{its two proposals, argmax and sampled. Both are run at block size $15$, so that every arm
drafts the same fifteen tokens and verifies the same sixteen slots}. \chg{\dflashtwo is likewise served under both, and at
its released selector width rather than at ours: matching the two candidate pools would change its
method rather than control a variable, since the two rerankers are different mechanisms at
different budgets, so what we match across arms is the drafted width instead. Each baseline is
composed from its own released branch rather than reimplemented from its published settings, and
\dflashtwo specifically is benchmarked through a third-party reimplementation trained under our
contract, since its own sources do not disclose a training recipe.} All
systems are trained on the same regenerated corpus for the same six epochs, at the same global
batch size and sequence length, \chg{with self-distillation disabled on every arm, and with
drafter and target sharing a rotary base of $10^{6}$}.

\begin{table}[t]
\centering
\small
\caption{Configuration of \method and of the joint training run. Symbols follow
Section~\ref{sec:method}. \chg{Every arm in the comparison is trained under this configuration,
with four keys changed for the smaller target. The step count follows from the conversation count
and the global batch size, $\lfloor N/64 \rfloor \times 6$, since batches are drawn over
conversations rather than tokens and the last partial batch is dropped. The sequence length is the
smallest common multiple of the arms' block sizes that reaches $3{,}072$.}
}
\label{tab:hparams}
\begin{tabular}{llc}
\toprule
Symbol & Quantity & Value \\
\midrule
$\block$   & Block size, of which drafted slots        & $16$, of which $15$ \\
$\topk$    & Candidates retained per slot              & $8$ \\
$d_e$      & Dimension of the in and out vectors       & $1024$ \\
$\sigma$   & Cosine scale                              & $8.0$ \\
           & Scoring Transformer layers                & $2$ \\
           & Hidden width, heads, feed-forward width   & $1024$, $8$, $2048$ \\
           & Parameters in the \method head            & $36.76$M (8B), $32.04$M (4B) \\
\midrule
$\beta$    & Decay of the \dflash cross-entropy        & $7.0$ \\
$\lambda$  & Weight of the distractor penalty          & $1.0$ \\
$\gamma$   & Weight of the reranker objective, Eq.~\eqref{eq:loss} & $0.25$ \\
\midrule
           & Optimizer                                 & AdamW \\
           & \chg{Learning rate}                       & \chg{$6\times10^{-4}$} \\
           & \chg{Weight decay}                        & \chg{$0$} \\
           & Schedule                                  & \chg{linear}, $4\%$ warmup \\
           & Gradient-norm clipping                    & $1.0$ \\
           & Global batch size, sequence length        & $64$, \chg{$3120$} \\
           & \chg{Anchor positions sampled per sequence} & \chg{$512$} \\
           & Epochs                                    & $6$ \\
           & \chg{Training conversations}              & \chg{$1{,}397{,}048$ (8B), $1{,}396{,}889$ (4B)} \\
           & \chg{Optimizer steps}                     & \chg{$130{,}968$ (8B), $130{,}956$ (4B)} \\
           & Precision                                 & bf16, fp32 master weights \\
           & Hardware                                  & \chg{$8$ nodes $\times\,8$ H100, DDP} \\
\bottomrule
\end{tabular}
\end{table}

\section{Additional Results}
\label{app:full-results}

\paragraph{Acceptance length and absolute throughput.}
Table~\ref{tab:main-merged} reports speedup, which is the quantity the comparison turns on.
Tables~\ref{tab:main-8b} and~\ref{tab:main-4b} give the two underlying measurements for each
target, the block-weighted acceptance length and the output throughput in tokens per second, under
the same protocol. They are the source of the acceptance-length statements of
Section~\ref{sec:exp-main}.

\begin{table}[!t]
\centering
\footnotesize
\setlength{\tabcolsep}{2pt}
\caption{\chg{Acceptance length ($\tau$, block-weighted) and throughput behind Table~\ref{tab:main-merged}, Qwen3-8B. Autoregressive decoding does not speculate and so has no $\tau$. The $T\!=\!1$ panel is the sampled proposal. Highest per row and temperature in bold, throughputs within the $1\%$ tie band of the fastest underlined.}}
\label{tab:main-8b}
\begin{tabular*}{\textwidth}{@{}c@{\hspace{5pt}}l@{\extracolsep{\fill}}cccccc@{}}
\toprule
& Benchmark & AR & \method & \dflashtwo & Domino & DSpark & \dflash \\
\midrule
\multirow{10}{*}{\rotatebox[origin=c]{90}{\scriptsize \textit{Greedy}, $T\!=\!0$}}
& GSM8K & 149 & \textbf{7.56} / \textbf{795} & 7.25 / 769 & 7.23 / 738 & 7.38 / 740 & 6.34 / 696 \\
& MATH-500 & 150 & \textbf{9.06} / \textbf{987} & 9.00 / \underline{984} & 8.98 / 946 & 9.01 / 932 & 7.91 / 891 \\
& AIME 2025 & 149 & \textbf{8.16} / \textbf{899} & 7.97 / 882 & 8.07 / 860 & 8.04 / 842 & 7.13 / 811 \\
& HumanEval & 146 & 7.08 / 588 & 7.08 / \textbf{600} & 6.86 / 568 & \textbf{7.16} / 582 & 6.16 / 553 \\
& MBPP & 150 & 5.85 / \textbf{626} & 5.68 / 614 & 5.68 / 591 & \textbf{5.89} / 601 & 5.03 / 560 \\
& LiveCodeBench & 148 & 7.75 / \textbf{804} & 7.60 / 792 & 7.55 / 759 & \textbf{7.78} / 768 & 6.81 / 733 \\
& AlpacaEval & 150 & \textbf{3.66} / \textbf{405} & 3.47 / 386 & 3.63 / 387 & 3.59 / 377 & 3.22 / 368 \\
& MT-Bench & 150 & 3.94 / \textbf{435} & 3.75 / 415 & 3.95 / 421 & \textbf{3.96} / 414 & 3.48 / 396 \\
& SPEED & 149 & \textbf{4.92} / \textbf{538} & 4.65 / 512 & 4.76 / 504 & 4.82 / 501 & 4.27 / 482 \\
\cmidrule(lr){2-8}
& Average & 149 & \textbf{6.44} / \textbf{675} & 6.27 / 662 & 6.30 / 641 & 6.40 / 640 & 5.59 / 610 \\
\addlinespace[4pt]
\multirow{10}{*}{\rotatebox[origin=c]{90}{\scriptsize \textit{Sampling}, $T\!=\!1$}}
& GSM8K & 148 & 6.86 / \textbf{719} & 6.51 / 574 & 6.22 / 636 & \textbf{6.87} / 669 & 5.58 / 609 \\
& MATH-500 & 148 & 6.83 / \underline{739} & 6.84 / 621 & 6.18 / 649 & \textbf{7.43} / \textbf{743} & 5.67 / 635 \\
& AIME 2025 & 147 & 5.38 / 588 & 5.45 / 498 & 4.76 / 505 & \textbf{6.06} / \textbf{613} & 4.54 / 512 \\
& HumanEval & 143 & \textbf{8.28} / \textbf{646} & 7.83 / 540 & 7.90 / 606 & 7.85 / 586 & 7.09 / 591 \\
& MBPP & 148 & 5.18 / \textbf{552} & 5.06 / 452 & 4.77 / 494 & \textbf{5.46} / 538 & 4.36 / 482 \\
& LiveCodeBench & 146 & 6.02 / \textbf{622} & 5.88 / 479 & 5.62 / 565 & \textbf{6.16} / 590 & 5.13 / 551 \\
& AlpacaEval & 148 & 3.19 / \textbf{350} & 3.07 / 282 & 2.90 / 309 & \textbf{3.27} / 332 & 2.68 / 304 \\
& MT-Bench & 148 & 3.63 / \textbf{395} & 3.44 / 315 & 3.35 / 354 & \textbf{3.69} / 372 & 3.00 / 338 \\
& SPEED & 147 & 3.75 / \textbf{407} & 3.62 / 329 & 3.37 / 355 & \textbf{3.85} / 387 & 3.09 / 347 \\
\cmidrule(lr){2-8}
& Average & 147 & 5.46 / \textbf{558} & 5.30 / 454 & 5.01 / 497 & \textbf{5.63} / 537 & 4.57 / 486 \\
\bottomrule
\end{tabular*}
\end{table}

\begin{table}[!t]
\centering
\footnotesize
\setlength{\tabcolsep}{2pt}
\caption{\chg{As Table~\ref{tab:main-8b}, Qwen3-4B.}}
\label{tab:main-4b}
\begin{tabular*}{\textwidth}{@{}c@{\hspace{5pt}}l@{\extracolsep{\fill}}cccccc@{}}
\toprule
& Benchmark & AR & \method & \dflashtwo & Domino & DSpark & \dflash \\
\midrule
\multirow{10}{*}{\rotatebox[origin=c]{90}{\scriptsize \textit{Greedy}, $T\!=\!0$}}
& GSM8K & 234 & \textbf{7.50} / \textbf{1171} & 7.27 / 1147 & 7.15 / 1068 & 7.24 / 1054 & 6.28 / 1038 \\
& MATH-500 & 235 & \textbf{8.88} / \underline{1438} & 8.85 / \textbf{1441} & 8.80 / 1356 & 8.70 / 1304 & 7.74 / 1314 \\
& AIME 2025 & 233 & \textbf{8.35} / \underline{1375} & 8.32 / \textbf{1376} & 8.25 / 1289 & 8.14 / 1237 & 7.29 / 1253 \\
& HumanEval & 226 & 7.22 / 882 & \textbf{7.44} / \textbf{902} & 7.07 / 842 & 7.26 / 829 & 6.37 / 822 \\
& MBPP & 235 & \textbf{5.78} / \textbf{920} & 5.63 / 902 & 5.63 / 854 & 5.69 / 842 & 4.91 / 823 \\
& LiveCodeBench & 218 & \textbf{7.86} / \underline{1202} & 7.76 / \textbf{1213} & 7.59 / 1126 & 7.79 / 1123 & 6.94 / 1095 \\
& AlpacaEval & 236 & \textbf{3.63} / \textbf{609} & 3.47 / 587 & 3.62 / 577 & 3.59 / 556 & 3.22 / 565 \\
& MT-Bench & 234 & \textbf{3.99} / \textbf{655} & 3.80 / 630 & 3.93 / 614 & 3.93 / 598 & 3.51 / 604 \\
& SPEED & 232 & \textbf{4.99} / \textbf{812} & 4.81 / 789 & 4.86 / 753 & 4.86 / 733 & 4.38 / 746 \\
\cmidrule(lr){2-8}
& Average & 232 & \textbf{6.47} / \textbf{1007} & 6.37 / \underline{999} & 6.32 / 942 & 6.36 / 919 & 5.63 / 918 \\
\addlinespace[4pt]
\multirow{10}{*}{\rotatebox[origin=c]{90}{\scriptsize \textit{Sampling}, $T\!=\!1$}}
& GSM8K & 230 & \textbf{6.88} / \textbf{1063} & 6.57 / 793 & 6.35 / 943 & 6.79 / 939 & 5.62 / 921 \\
& MATH-500 & 228 & 6.89 / \underline{1024} & 6.92 / 850 & 6.26 / 951 & \textbf{7.32} / \textbf{1032} & 5.68 / 928 \\
& AIME 2025 & 228 & 5.52 / \textbf{897} & 5.45 / 683 & 4.84 / 750 & \textbf{5.95} / 860 & 4.41 / 748 \\
& HumanEval & 221 & 8.03 / \textbf{908} & \textbf{8.04} / 732 & 7.66 / 857 & 7.99 / 837 & 7.15 / 877 \\
& MBPP & 231 & \textbf{5.40} / \textbf{849} & 5.17 / 632 & 5.05 / 760 & 5.37 / 755 & 4.51 / 748 \\
& LiveCodeBench & 226 & \textbf{5.88} / \textbf{885} & 5.61 / 652 & 5.49 / 805 & 5.81 / 792 & 4.96 / 793 \\
& AlpacaEval & 229 & \textbf{3.32} / \textbf{526} & 3.11 / 374 & 3.13 / 487 & \textbf{3.32} / 470 & 2.83 / 484 \\
& MT-Bench & 230 & \textbf{3.67} / \textbf{596} & 3.41 / 427 & 3.48 / 539 & \textbf{3.67} / 531 & 3.15 / 537 \\
& SPEED & 228 & \textbf{3.98} / \textbf{642} & 3.77 / 471 & 3.62 / 549 & 3.97 / 572 & 3.33 / 548 \\
\cmidrule(lr){2-8}
& Average & 228 & 5.51 / \textbf{821} & 5.34 / 624 & 5.10 / 738 & \textbf{5.58} / 754 & 4.63 / 732 \\
\bottomrule
\end{tabular*}
\end{table}

\begin{table}[!t]
\centering
\footnotesize
\setlength{\tabcolsep}{2pt}
\caption{\chg{The argmax-proposal panel at $T\!=\!1$, the other half of the measurement whose sampled
panel is in Tables~\ref{tab:main-8b} and~\ref{tab:main-4b}. Reported as in
Table~\ref{tab:main-8b}.}}
\label{tab:main-greedy}
\begin{tabular*}{\textwidth}{@{}c@{\hspace{5pt}}l@{\extracolsep{\fill}}cccccc@{}}
\toprule
& Benchmark & AR & \method & \dflashtwo & Domino & DSpark & \dflash \\
\midrule
\multirow{10}{*}{\rotatebox[origin=c]{90}{\scriptsize \textit{Qwen3-8B}}}
& GSM8K & 148 & \textbf{6.49} / \textbf{685} & 6.25 / 549 & 6.22 / 636 & 6.32 / 640 & 5.58 / 609 \\
& MATH-500 & 148 & 6.23 / \textbf{678} & \textbf{6.28} / 568 & 6.18 / 649 & 6.22 / 646 & 5.67 / 635 \\
& AIME 2025 & 147 & \textbf{4.86} / \textbf{533} & 4.73 / 432 & 4.76 / 505 & 4.74 / 497 & 4.54 / 512 \\
& HumanEval & 143 & \textbf{8.33} / \textbf{649} & 8.13 / 547 & 7.90 / 606 & 8.02 / 611 & 7.09 / 591 \\
& MBPP & 148 & \textbf{4.96} / \textbf{530} & 4.85 / 431 & 4.77 / 494 & 4.92 / 503 & 4.36 / 482 \\
& LiveCodeBench & 146 & \textbf{5.87} / \textbf{607} & 5.77 / 499 & 5.62 / 565 & 5.82 / 558 & 5.13 / 551 \\
& AlpacaEval & 148 & \textbf{2.93} / \textbf{323} & 2.83 / 258 & 2.90 / 309 & 2.92 / 307 & 2.68 / 304 \\
& MT-Bench & 148 & 3.33 / \textbf{365} & 3.23 / 294 & \textbf{3.35} / 354 & 3.34 / 349 & 3.00 / 338 \\
& SPEED & 147 & \textbf{3.43} / \textbf{373} & 3.31 / 300 & 3.37 / 355 & 3.39 / 352 & 3.09 / 347 \\
\cmidrule(lr){2-8}
& Average & 147 & \textbf{5.16} / \textbf{527} & 5.04 / 431 & 5.01 / 497 & 5.08 / 496 & 4.57 / 486 \\
\addlinespace[4pt]
\multirow{10}{*}{\rotatebox[origin=c]{90}{\scriptsize \textit{Qwen3-4B}}}
& GSM8K & 230 & \textbf{6.57} / \textbf{1021} & 6.44 / 774 & 6.35 / 943 & 6.42 / 935 & 5.62 / 921 \\
& MATH-500 & 228 & \textbf{6.38} / \textbf{976} & 6.34 / 775 & 6.26 / 951 & 6.31 / 929 & 5.68 / 928 \\
& AIME 2025 & 228 & 4.81 / \textbf{783} & \textbf{4.87} / 608 & 4.84 / 750 & 4.75 / 722 & 4.41 / 748 \\
& HumanEval & 221 & 8.20 / \textbf{929} & \textbf{8.50} / 758 & 7.66 / 857 & 8.09 / 868 & 7.15 / 877 \\
& MBPP & 231 & \textbf{5.23} / \textbf{823} & 5.09 / 618 & 5.05 / 760 & 5.19 / 764 & 4.51 / 748 \\
& LiveCodeBench & 226 & \textbf{5.61} / \textbf{861} & 5.53 / 648 & 5.49 / 805 & 5.54 / 774 & 4.96 / 793 \\
& AlpacaEval & 229 & \textbf{3.16} / \textbf{510} & 3.02 / 376 & 3.13 / 487 & 3.10 / 451 & 2.83 / 484 \\
& MT-Bench & 230 & 3.46 / \textbf{562} & 3.35 / 417 & \textbf{3.48} / 539 & 3.45 / 524 & 3.15 / 537 \\
& SPEED & 228 & \textbf{3.70} / \textbf{599} & 3.60 / 443 & 3.62 / 549 & 3.66 / 553 & 3.33 / 548 \\
\cmidrule(lr){2-8}
& Average & 228 & \textbf{5.24} / \textbf{785} & 5.20 / 602 & 5.10 / 738 & 5.17 / 724 & 4.63 / 732 \\
\bottomrule
\end{tabular*}
\end{table}

\paragraph{Per-category results on SPEED.}
Tables~\ref{tab:speed-8b} and \ref{tab:speed-4b} break the SPEED aggregate of
Section~\ref{sec:exp-main} into its eleven categories, under the same protocol. \chg{The per-benchmark
conclusions carry over: \method serves faster than every system compared in all forty-four
cells. The multilingual
lead noted in Section~\ref{sec:exp-main} is the clearest of any category on both axes, reaching
$7.4\%$ in acceptance length and $6.3\%$ in throughput on Qwen3-4B under greedy decoding. Sample
standard deviations across the three $T\!=\!1$ seeds stay below $0.28$ on acceptance length in every
category and system, with one exception, roleplay on Qwen3-4B.}

\paragraph{Seed variability at $T\!=\!1$.}
Tables~\ref{tab:seed-8b} and \ref{tab:seed-4b} give the mean and sample standard deviation over
seeds $42$, $43$, and $44$ that stand behind the $T\!=\!1$ panels of Tables~\ref{tab:main-8b}
and~\ref{tab:main-4b}. \chg{Dispersion is small throughout: the standard deviation of $\tau$ stays below
$0.25$ on every benchmark and system, and below $0.16$ on all but eight of the ninety
cells. What scatter remains
concentrates on the small splits, above all AIME 2025 and HumanEval, which at $30$ and $164$ prompts
are the two smallest in the suite. On the two splits of more than a thousand prompts no standard
deviation exceeds $0.10$.}

\begin{table}[t]
\centering
\small
\setlength{\tabcolsep}{3pt}
\caption{\chg{Per-category results on SPEED \citep{abramovich2026speed}, Qwen3-8B, reported as in Table~\ref{tab:main-8b}.}}
\label{tab:speed-8b}
\begin{tabular}{c@{\hspace{5pt}}l ccccc}
\toprule
& Category & \method & \dflashtwo & Domino & DSpark & \dflash \\
\midrule
\multirow{11}{*}{\rotatebox[origin=c]{90}{\scriptsize \textit{Greedy}, $T\!=\!0$}}
& Math & \textbf{7.01} / \textbf{772} & 6.84 / 757 & 6.75 / 719 & 6.92 / 724 & 6.05 / 687 \\
& Coding & \textbf{6.51} / \textbf{707} & 6.09 / 665 & 6.14 / 645 & 6.30 / 650 & 5.43 / 609 \\
& Multilingual & \textbf{5.54} / \textbf{607} & 4.98 / 550 & 4.89 / 521 & 5.15 / 536 & 4.59 / 519 \\
& STEM & \textbf{6.21} / \textbf{686} & 5.89 / 654 & 6.03 / 646 & 6.10 / 641 & 5.42 / 619 \\
& Humanities & \textbf{5.74} / \textbf{634} & 5.43 / 603 & 5.67 / 605 & 5.67 / 595 & 5.03 / 572 \\
& Reasoning & \textbf{5.09} / \textbf{553} & 4.85 / 530 & 4.99 / 524 & 5.00 / 517 & 4.39 / 493 \\
& RAG & \textbf{4.72} / \textbf{493} & 4.45 / 470 & 4.52 / 460 & 4.57 / 456 & 4.02 / 438 \\
& Writing & \textbf{3.74} / \textbf{413} & 3.52 / 392 & 3.65 / 391 & 3.66 / 385 & 3.25 / 370 \\
& QA & \textbf{3.43} / \textbf{378} & 3.27 / 362 & 3.37 / 360 & 3.42 / 359 & 3.02 / 344 \\
& Summarization & \textbf{3.25} / \textbf{345} & 3.01 / 323 & 3.01 / 312 & 3.12 / 316 & 2.80 / 308 \\
& Roleplay & \textbf{2.88} / \textbf{313} & 2.74 / 300 & 2.84 / 300 & 2.85 / 295 & 2.63 / 296 \\
\addlinespace[4pt]
\multirow{11}{*}{\rotatebox[origin=c]{90}{\scriptsize \textit{Sampling}, $T\!=\!1$}}
& Math & 4.64 / \textbf{508} & 4.65 / 425 & 3.99 / 424 & \textbf{4.99} / \underline{504} & 3.77 / 426 \\
& Coding & 5.11 / \textbf{550} & 4.85 / 438 & 4.56 / 476 & \textbf{5.28} / 526 & 4.12 / 460 \\
& Multilingual & \textbf{3.88} / \textbf{422} & 3.56 / 326 & 3.56 / 375 & 3.62 / 365 & 3.27 / 367 \\
& STEM & 3.93 / \textbf{431} & 3.88 / 356 & 3.45 / 366 & \textbf{4.15} / 421 & 3.17 / 360 \\
& Humanities & 3.68 / \textbf{402} & 3.53 / 323 & 3.25 / 344 & \textbf{3.83} / 387 & 3.01 / 340 \\
& Reasoning & 4.26 / \textbf{459} & 4.11 / 371 & 3.88 / 406 & \textbf{4.32} / 431 & 3.48 / 388 \\
& RAG & 4.23 / \textbf{441} & 4.02 / 353 & 3.80 / 386 & \textbf{4.25} / 411 & 3.57 / 387 \\
& Writing & 3.12 / \textbf{341} & 3.02 / 275 & 2.81 / 298 & \textbf{3.19} / 323 & 2.57 / 290 \\
& QA & 2.95 / \textbf{321} & 2.84 / 260 & 2.75 / 291 & \textbf{3.06} / 308 & 2.51 / 283 \\
& Summarization & \textbf{3.20} / \textbf{334} & 2.93 / 260 & 2.89 / 296 & 3.13 / 306 & 2.74 / 298 \\
& Roleplay & \textbf{2.50} / \textbf{270} & 2.34 / 212 & 2.39 / 249 & \textbf{2.50} / 250 & 2.17 / 242 \\
\bottomrule
\end{tabular}
\end{table}

\begin{table}[t]
\centering
\small
\setlength{\tabcolsep}{3pt}
\caption{\chg{As Table~\ref{tab:speed-8b}, Qwen3-4B.}}
\label{tab:speed-4b}
\begin{tabular}{c@{\hspace{5pt}}l ccccc}
\toprule
& Category & \method & \dflashtwo & Domino & DSpark & \dflash \\
\midrule
\multirow{11}{*}{\rotatebox[origin=c]{90}{\scriptsize \textit{Greedy}, $T\!=\!0$}}
& Math & \textbf{7.41} / \textbf{1216} & 7.22 / 1197 & 7.25 / 1137 & 7.14 / 1087 & 6.46 / 1111 \\
& Coding & \textbf{6.38} / \textbf{1029} & 6.25 / \underline{1019} & 6.15 / 949 & 6.26 / 940 & 5.46 / 926 \\
& Multilingual & \textbf{5.83} / \textbf{948} & 5.43 / 891 & 5.17 / 801 & 5.27 / 794 & 4.75 / 810 \\
& STEM & \textbf{5.75} / \textbf{950} & 5.51 / 918 & 5.51 / 868 & 5.57 / 852 & 5.00 / 863 \\
& Humanities & \textbf{5.33} / \textbf{878} & 5.18 / 860 & 5.32 / 833 & 5.18 / 790 & 4.72 / 813 \\
& Reasoning & \textbf{5.37} / \textbf{868} & 5.25 / 855 & 5.28 / 813 & 5.29 / 793 & 4.71 / 798 \\
& RAG & \textbf{4.47} / \textbf{702} & 4.41 / \underline{697} & 4.23 / 634 & 4.41 / 645 & 3.91 / 645 \\
& Writing & \textbf{3.69} / \textbf{609} & 3.52 / 585 & 3.65 / 572 & 3.65 / 556 & 3.27 / 565 \\
& QA & 3.55 / \textbf{584} & 3.44 / 572 & \textbf{3.59} / 561 & 3.55 / 541 & 3.16 / 546 \\
& Summarization & \textbf{3.48} / \textbf{549} & 3.25 / 519 & 3.21 / 483 & 3.27 / 480 & 3.00 / 497 \\
& Roleplay & \textbf{3.44} / \textbf{557} & 3.33 / 542 & 3.42 / 525 & 3.39 / 507 & 3.14 / 531 \\
\addlinespace[4pt]
\multirow{11}{*}{\rotatebox[origin=c]{90}{\scriptsize \textit{Sampling}, $T\!=\!1$}}
& Math & 4.73 / \textbf{772} & 4.59 / 579 & 4.19 / 651 & \textbf{4.97} / 723 & 3.87 / 660 \\
& Coding & 5.20 / \textbf{826} & 5.06 / 626 & 4.83 / 734 & \textbf{5.25} / 747 & 4.28 / 715 \\
& Multilingual & \textbf{4.17} / \textbf{671} & 3.67 / 457 & 3.52 / 510 & 3.79 / 544 & 3.41 / 521 \\
& STEM & 4.17 / \textbf{684} & 4.07 / 516 & 3.69 / 577 & \textbf{4.26} / 624 & 3.42 / 587 \\
& Humanities & 3.85 / \textbf{627} & 3.64 / 459 & 3.45 / 536 & \textbf{3.89} / 565 & 3.24 / 552 \\
& Reasoning & 4.48 / \textbf{719} & 4.29 / 535 & 4.17 / 640 & \textbf{4.50} / 645 & 3.71 / 626 \\
& RAG & \textbf{4.42} / \textbf{684} & 4.20 / 507 & 3.91 / 582 & 4.32 / 600 & 3.78 / 615 \\
& Writing & \textbf{3.36} / \textbf{546} & 3.15 / 396 & 3.14 / 487 & 3.33 / 484 & 2.83 / 481 \\
& QA & 3.23 / \textbf{523} & 3.07 / 385 & 3.08 / 473 & \textbf{3.27} / 473 & 2.77 / 471 \\
& Summarization & \textbf{3.39} / \textbf{523} & 3.09 / 372 & 3.13 / 434 & 3.21 / 446 & 2.92 / 382 \\
& Roleplay & \textbf{2.84} / \textbf{452} & 2.52 / 311 & 2.54 / 358 & 2.64 / 376 & 2.39 / 370 \\
\bottomrule
\end{tabular}
\end{table}

\begin{table}[t]
\centering
\small
\setlength{\tabcolsep}{3.5pt}
\caption{\chg{Seed variability behind the $T\!=\!1$ panel of Table~\ref{tab:main-8b}. Mean over seeds $42$, $43$ and $44$, sample standard deviation in parentheses.}}
\label{tab:seed-8b}
\begin{tabular}{c@{\hspace{5pt}}l ccccc}
\toprule
& Benchmark & \method & \dflashtwo & Domino & DSpark & \dflash \\
\midrule
\multirow{9}{*}{\rotatebox[origin=c]{90}{\scriptsize \textit{Acceptance} $\tau$}}
& GSM8K & 6.86 (0.04) & 6.51 (0.03) & 6.22 (0.01) & 6.87 (0.03) & 5.58 (0.02) \\
& MATH-500 & 6.83 (0.03) & 6.84 (0.04) & 6.18 (0.02) & 7.43 (0.07) & 5.67 (0.05) \\
& AIME 2025 & 5.38 (0.04) & 5.45 (0.05) & 4.76 (0.13) & 6.06 (0.17) & 4.54 (0.20) \\
& HumanEval & 8.28 (0.21) & 7.83 (0.19) & 7.90 (0.05) & 7.85 (0.10) & 7.09 (0.09) \\
& MBPP & 5.18 (0.03) & 5.06 (0.01) & 4.77 (0.03) & 5.46 (0.04) & 4.36 (0.05) \\
& LiveCodeBench & 6.02 (0.07) & 5.88 (0.04) & 5.62 (0.00) & 6.16 (0.09) & 5.13 (0.05) \\
& AlpacaEval & 3.19 (0.01) & 3.07 (0.02) & 2.90 (0.01) & 3.27 (0.01) & 2.68 (0.00) \\
& MT-Bench & 3.63 (0.01) & 3.44 (0.03) & 3.35 (0.03) & 3.69 (0.02) & 3.00 (0.03) \\
& SPEED & 3.75 (0.02) & 3.62 (0.02) & 3.37 (0.03) & 3.85 (0.01) & 3.09 (0.01) \\
\addlinespace[4pt]
\multirow{9}{*}{\rotatebox[origin=c]{90}{\scriptsize \textit{Throughput}}}
& GSM8K & 719 (4.4) & 574 (1.8) & 636 (1.2) & 669 (2.6) & 609 (2.4) \\
& MATH-500 & 739 (2.2) & 621 (3.6) & 649 (3.1) & 743 (6.1) & 635 (4.3) \\
& AIME 2025 & 588 (4.7) & 498 (3.7) & 505 (13.8) & 613 (16.6) & 512 (22.3) \\
& HumanEval & 646 (9.4) & 540 (11.0) & 606 (3.5) & 586 (4.5) & 591 (2.7) \\
& MBPP & 552 (2.6) & 452 (1.9) & 494 (3.4) & 538 (4.1) & 482 (5.7) \\
& LiveCodeBench & 622 (6.0) & 479 (57.0) & 565 (0.3) & 590 (7.5) & 551 (4.9) \\
& AlpacaEval & 350 (1.3) & 282 (1.2) & 309 (0.9) & 332 (1.0) & 304 (0.5) \\
& MT-Bench & 395 (1.9) & 315 (3.5) & 354 (4.7) & 372 (3.1) & 338 (4.9) \\
& SPEED & 407 (1.8) & 329 (1.5) & 355 (2.9) & 387 (1.4) & 347 (1.8) \\
\bottomrule
\end{tabular}
\end{table}

\begin{table}[t]
\centering
\small
\setlength{\tabcolsep}{3.5pt}
\caption{\chg{As Table~\ref{tab:seed-8b}, Qwen3-4B.}}
\label{tab:seed-4b}
\begin{tabular}{c@{\hspace{5pt}}l ccccc}
\toprule
& Benchmark & \method & \dflashtwo & Domino & DSpark & \dflash \\
\midrule
\multirow{9}{*}{\rotatebox[origin=c]{90}{\scriptsize \textit{Acceptance} $\tau$}}
& GSM8K & 6.88 (0.01) & 6.57 (0.01) & 6.35 (0.01) & 6.79 (0.04) & 5.62 (0.02) \\
& MATH-500 & 6.89 (0.06) & 6.92 (0.05) & 6.26 (0.06) & 7.32 (0.04) & 5.68 (0.02) \\
& AIME 2025 & 5.52 (0.15) & 5.45 (0.18) & 4.84 (0.08) & 5.95 (0.10) & 4.41 (0.11) \\
& HumanEval & 8.03 (0.14) & 8.04 (0.24) & 7.66 (0.07) & 7.99 (0.18) & 7.15 (0.20) \\
& MBPP & 5.40 (0.05) & 5.17 (0.05) & 5.05 (0.02) & 5.37 (0.03) & 4.51 (0.06) \\
& LiveCodeBench & 5.88 (0.01) & 5.61 (0.01) & 5.49 (0.05) & 5.81 (0.07) & 4.96 (0.06) \\
& AlpacaEval & 3.32 (0.00) & 3.11 (0.01) & 3.13 (0.00) & 3.32 (0.00) & 2.83 (0.00) \\
& MT-Bench & 3.67 (0.03) & 3.41 (0.03) & 3.48 (0.06) & 3.67 (0.04) & 3.15 (0.04) \\
& SPEED & 3.98 (0.06) & 3.77 (0.02) & 3.62 (0.02) & 3.97 (0.03) & 3.33 (0.04) \\
\addlinespace[4pt]
\multirow{9}{*}{\rotatebox[origin=c]{90}{\scriptsize \textit{Throughput}}}
& GSM8K & 1063 (3.9) & 793 (1.3) & 943 (4.6) & 939 (6.0) & 921 (2.9) \\
& MATH-500 & 1024 (133.9) & 850 (22.0) & 951 (10.5) & 1032 (31.2) & 928 (48.0) \\
& AIME 2025 & 897 (24.6) & 683 (23.3) & 750 (11.7) & 860 (13.8) & 748 (18.2) \\
& HumanEval & 908 (5.5) & 732 (15.8) & 857 (4.4) & 837 (9.8) & 877 (14.8) \\
& MBPP & 849 (12.2) & 632 (8.1) & 760 (6.1) & 755 (8.3) & 748 (12.0) \\
& LiveCodeBench & 885 (19.1) & 652 (21.3) & 805 (5.3) & 792 (3.0) & 793 (16.9) \\
& AlpacaEval & 526 (27.2) & 374 (29.4) & 487 (0.3) & 470 (14.7) & 484 (0.1) \\
& MT-Bench & 596 (8.8) & 427 (4.2) & 539 (11.3) & 531 (7.3) & 537 (9.4) \\
& SPEED & 642 (11.4) & 471 (1.7) & 549 (7.2) & 572 (4.4) & 548 (27.2) \\
\bottomrule
\end{tabular}
\end{table}

\chg{\section{The Convolution Variant}}
\label{app:conv}

\chg{\dflashtwo contributes a grouped causal convolution inside the draft layers alongside its
candidate selector. The two are independent: the convolution improves the drafter, and is
indifferent to how the candidates it proposes are afterwards correlated. We therefore place it
under \method as well and report the resulting arm, \methodc, here rather than in the main tables,
since the question it answers is not how \method compares with the baselines but how much a better
drafter adds on top of it.}

\chg{Table~\ref{tab:conv-merged} gives the nine benchmarks at both target sizes and both decoding
regimes. Its baseline columns are the same measurement as in Table~\ref{tab:main-merged}, so the
two tables may be read against one another. Table~\ref{tab:conv-tau} isolates what the convolution
itself buys. Under greedy decoding it raises acceptance length over \method by $2.3\%$ on Qwen3-8B
and $2.5\%$ on Qwen3-4B on the nine-benchmark average, and on every one of the eighteen benchmarks
individually. It converts that into throughput at the
larger target size, by $0.6\%$ on the average, but not at the smaller, where its fixed per-block
cost is a larger share of a cheaper block. Over the seventy-two settings of
Section~\ref{sec:exp-main} the arm holds the highest throughput at $64$, ties at $6$ and trails at
$2$, against $63$, $6$ and $3$ for \method, and it removes both HumanEval deficits: at Qwen3-4B it
accepts $5.4\%$ more than \dflashtwo where \method accepts $3.0\%$ less. On the concurrency sweep of
Section~\ref{sec:exp-concurrency} it leads all thirty-six operating points, as \method does, with an
acceptance margin over the strongest baseline in each cell of $2.2$ to $10.4\%$ against \method's
$1.1$ to $5.7\%$. Table~\ref{tab:conv-isl} places the two side by side there: the convolution
accepts more at every one of the thirty-six points, by $0.3$ to $5.2\%$, and serves faster at
twenty-six of them. The convolution is not free: it adds $0.136$ ms per block, raising the cost of the
\method head by about half (Table~\ref{tab:cost}).}

\begin{table}[!t]
\centering
\footnotesize
\setlength{\tabcolsep}{1.6pt}
\caption{\chg{\methodc against the same baselines, reported as in Table~\ref{tab:main-merged}. The
baseline columns are the same measurement, so the two tables read against one another.}}
\label{tab:conv-merged}
\begin{tabular*}{\textwidth}{@{}c@{\hspace{3pt}}l@{\extracolsep{\fill}}|cc|cc|cc|cc|cc|cc@{}}
\toprule
& & \multicolumn{2}{c|}{AR (TPS)} & \multicolumn{2}{c|}{\methodc} & \multicolumn{2}{c|}{\dflashtwo} & \multicolumn{2}{c|}{Domino} & \multicolumn{2}{c|}{DSpark} & \multicolumn{2}{c}{\dflash} \\
& Benchmark & 8B & 4B & 8B & 4B & 8B & 4B & 8B & 4B & 8B & 4B & 8B & 4B \\
\midrule
\multirow{10}{*}{\rotatebox[origin=c]{90}{\scriptsize \textit{Greedy}, $T\!=\!0$}}
& GSM8K & 149 & 234 & \textbf{5.35}$\times$ & \textbf{4.99}$\times$ & 5.15$\times$ & 4.90$\times$ & 4.94$\times$ & 4.56$\times$ & 4.95$\times$ & 4.50$\times$ & 4.66$\times$ & 4.43$\times$ \\
& MATH-500 & 150 & 235 & \textbf{6.59}$\times$ & \underline{6.12}$\times$ & \underline{6.54}$\times$ & \textbf{6.14}$\times$ & 6.29$\times$ & 5.78$\times$ & 6.20$\times$ & 5.55$\times$ & 5.93$\times$ & 5.60$\times$ \\
& AIME 2025 & 149 & 233 & \textbf{6.02}$\times$ & \underline{5.89}$\times$ & 5.90$\times$ & \textbf{5.90}$\times$ & 5.75$\times$ & 5.53$\times$ & 5.64$\times$ & 5.30$\times$ & 5.42$\times$ & 5.37$\times$ \\
& HumanEval & 146 & 226 & \textbf{4.13}$\times$ & \textbf{4.08}$\times$ & \underline{4.11}$\times$ & 3.99$\times$ & 3.89$\times$ & 3.73$\times$ & 3.99$\times$ & 3.67$\times$ & 3.79$\times$ & 3.64$\times$ \\
& MBPP & 150 & 235 & \textbf{4.21}$\times$ & \underline{3.80}$\times$ & 4.08$\times$ & \textbf{3.83}$\times$ & 3.93$\times$ & 3.63$\times$ & 3.99$\times$ & 3.58$\times$ & 3.72$\times$ & 3.50$\times$ \\
& LiveCodeBench & 148 & 218 & \textbf{5.48}$\times$ & \underline{5.57}$\times$ & 5.35$\times$ & \textbf{5.58}$\times$ & 5.12$\times$ & 5.18$\times$ & 5.18$\times$ & 5.16$\times$ & 4.94$\times$ & 5.03$\times$ \\
& AlpacaEval & 150 & 236 & \textbf{2.69}$\times$ & \textbf{2.53}$\times$ & 2.58$\times$ & 2.48$\times$ & 2.59$\times$ & 2.44$\times$ & 2.52$\times$ & 2.35$\times$ & 2.46$\times$ & 2.39$\times$ \\
& MT-Bench & 150 & 234 & \textbf{2.91}$\times$ & \textbf{2.78}$\times$ & 2.77$\times$ & 2.69$\times$ & 2.81$\times$ & 2.62$\times$ & 2.76$\times$ & 2.55$\times$ & 2.64$\times$ & 2.58$\times$ \\
& SPEED & 149 & 232 & \textbf{3.61}$\times$ & \textbf{3.52}$\times$ & 3.44$\times$ & 3.40$\times$ & 3.38$\times$ & 3.25$\times$ & 3.37$\times$ & 3.16$\times$ & 3.24$\times$ & 3.22$\times$ \\
\cmidrule(lr){2-14}
& Average & 149 & 232 & \textbf{4.56}$\times$ & \textbf{4.36}$\times$ & 4.44$\times$ & \underline{4.32}$\times$ & 4.30$\times$ & 4.08$\times$ & 4.29$\times$ & 3.98$\times$ & 4.09$\times$ & 3.97$\times$ \\
\addlinespace[4pt]
\multirow{10}{*}{\rotatebox[origin=c]{90}{\scriptsize \textit{Sampling}, $T\!=\!1$}}
& GSM8K & 148 & 230 & \textbf{4.82}$\times$ & \textbf{4.61}$\times$ & 3.88$\times$ & 3.45$\times$ & 4.31$\times$ & 4.10$\times$ & 4.53$\times$ & 4.09$\times$ & 4.13$\times$ & 4.01$\times$ \\
& MATH-500 & 148 & 228 & 4.95$\times$ & \textbf{4.86}$\times$ & 4.20$\times$ & 3.72$\times$ & 4.40$\times$ & 4.17$\times$ & \textbf{5.03}$\times$ & 4.52$\times$ & 4.30$\times$ & 4.07$\times$ \\
& AIME 2025 & 147 & 228 & 3.99$\times$ & \textbf{3.97}$\times$ & 3.38$\times$ & 2.99$\times$ & 3.43$\times$ & 3.29$\times$ & \textbf{4.16}$\times$ & 3.77$\times$ & 3.47$\times$ & 3.28$\times$ \\
& HumanEval & 143 & 221 & \textbf{4.56}$\times$ & \textbf{4.32}$\times$ & 3.77$\times$ & 3.32$\times$ & 4.24$\times$ & 3.89$\times$ & 4.09$\times$ & 3.80$\times$ & 4.13$\times$ & 3.97$\times$ \\
& MBPP & 148 & 231 & \textbf{3.73}$\times$ & \textbf{3.63}$\times$ & 3.06$\times$ & 2.74$\times$ & 3.34$\times$ & 3.29$\times$ & 3.64$\times$ & 3.27$\times$ & 3.26$\times$ & 3.24$\times$ \\
& LiveCodeBench & 146 & 226 & \textbf{4.33}$\times$ & \textbf{3.76}$\times$ & 3.29$\times$ & 2.88$\times$ & 3.88$\times$ & 3.56$\times$ & 4.05$\times$ & 3.50$\times$ & 3.78$\times$ & 3.51$\times$ \\
& AlpacaEval & 148 & 229 & \textbf{2.36}$\times$ & \textbf{2.35}$\times$ & 1.90$\times$ & 1.64$\times$ & 2.08$\times$ & 2.13$\times$ & 2.24$\times$ & 2.06$\times$ & 2.05$\times$ & 2.11$\times$ \\
& MT-Bench & 148 & 230 & \textbf{2.66}$\times$ & \textbf{2.58}$\times$ & 2.13$\times$ & 1.85$\times$ & 2.39$\times$ & 2.34$\times$ & 2.51$\times$ & 2.31$\times$ & 2.29$\times$ & 2.33$\times$ \\
& SPEED & 147 & 228 & \textbf{2.77}$\times$ & \textbf{2.78}$\times$ & 2.23$\times$ & 2.06$\times$ & 2.41$\times$ & 2.41$\times$ & 2.63$\times$ & 2.50$\times$ & 2.36$\times$ & 2.40$\times$ \\
\cmidrule(lr){2-14}
& Average & 147 & 228 & \textbf{3.80}$\times$ & \textbf{3.65}$\times$ & 3.10$\times$ & 2.74$\times$ & 3.39$\times$ & 3.24$\times$ & 3.65$\times$ & 3.31$\times$ & 3.31$\times$ & 3.21$\times$ \\
\bottomrule
\end{tabular*}
\end{table}

\begin{table}[t]
\centering
\footnotesize
\setlength{\tabcolsep}{3pt}
\caption{\chg{What the convolution buys, under greedy decoding. The two models differ only in the
grouped causal convolution inside the draft layers, written $+$\,conv here. Parentheses give the
percentage change from \method, and the better of each pair is in bold.}}
\label{tab:conv-tau}
\begin{tabular*}{\textwidth}{@{}l@{\extracolsep{\fill}}cccc cccc@{}}
\toprule
& \multicolumn{4}{c}{Qwen3-8B} & \multicolumn{4}{c}{Qwen3-4B} \\
\cmidrule(lr){2-5}\cmidrule(lr){6-9}
& \multicolumn{2}{c}{$\tau$} & \multicolumn{2}{c}{TPS} & \multicolumn{2}{c}{$\tau$} & \multicolumn{2}{c}{TPS} \\
\cmidrule(lr){2-3}\cmidrule(lr){4-5}\cmidrule(lr){6-7}\cmidrule(lr){8-9}
& \method & $+$\,conv & \method & $+$\,conv & \method & $+$\,conv & \method & $+$\,conv \\
\midrule
GSM8K & 7.56 & \textbf{7.72 (+2.1)} & 795 & \textbf{799 (+0.5)} & 7.50 & \textbf{7.63 (+1.8)} & \textbf{1171} & 1170 (-0.1) \\
MATH & 9.06 & \textbf{9.24 (+2.0)} & 987 & \textbf{991 (+0.4)} & 8.88 & \textbf{9.04 (+1.9)} & \textbf{1438} & 1437 (-0.0) \\
AIME & 8.16 & \textbf{8.28 (+1.6)} & 899 & \textbf{900 (+0.1)} & 8.35 & \textbf{8.52 (+2.0)} & 1375 & \textbf{1375 (+0.0)} \\
HEval & 7.08 & \textbf{7.39 (+4.5)} & 588 & \textbf{603 (+2.7)} & 7.22 & \textbf{7.84 (+8.7)} & 882 & \textbf{922 (+4.5)} \\
MBPP & 5.85 & \textbf{6.00 (+2.6)} & 626 & \textbf{633 (+1.0)} & 5.78 & \textbf{5.87 (+1.5)} & \textbf{920} & 894 (-2.8) \\
LCB & 7.75 & \textbf{7.97 (+2.8)} & 804 & \textbf{812 (+1.1)} & 7.86 & \textbf{7.95 (+1.1)} & 1202 & \textbf{1211 (+0.8)} \\
Alpaca & 3.66 & \textbf{3.70 (+1.1)} & \textbf{405} & 403 (-0.4) & 3.63 & \textbf{3.68 (+1.4)} & \textbf{609} & 597 (-1.9) \\
MTB & 3.94 & \textbf{4.01 (+1.9)} & 435 & \textbf{436 (+0.2)} & 3.99 & \textbf{4.03 (+1.1)} & \textbf{655} & 651 (-0.7) \\
SPEED & 4.92 & \textbf{4.99 (+1.4)} & 538 & \textbf{538 (-0.1)} & 4.99 & \textbf{5.11 (+2.4)} & 812 & \textbf{816 (+0.5)} \\
\midrule
Average & 6.44 & \textbf{6.59 (+2.3)} & 675 & \textbf{679 (+0.6)} & 6.47 & \textbf{6.63 (+2.5)} & 1007 & \textbf{1008 (+0.1)} \\
\bottomrule
\end{tabular*}
\end{table}

\begin{table}[t]
\centering
\footnotesize
\setlength{\tabcolsep}{2pt}
\caption{\chg{The convolution on the concurrency sweep, Qwen3-8B, greedy. Acceptance length and
throughput at the operating points of Table~\ref{tab:concurrency}, for \method and \methodc. Higher
of each pair in bold.}}
\label{tab:conv-isl}
\chg{\begin{tabular*}{\textwidth}{@{}c@{\hspace{3pt}}c@{\extracolsep{\fill}}|cc|cc|cc@{}}
\toprule
& & \multicolumn{2}{c|}{Low} & \multicolumn{2}{c|}{Mixed} & \multicolumn{2}{c}{High} \\
& $c$ & \method & $+$\,conv & \method & $+$\,conv & \method & $+$\,conv \\
\midrule
\multirow{6}{*}{\rotatebox[origin=c]{90}{\scriptsize \textit{$1K$}}}
& 1 & 4.89 / 515 & \textbf{5.10} / \textbf{528} & 4.91 / 526 & \textbf{5.00} / \textbf{527} & 3.11 / \textbf{343} & \textbf{3.15} / 342 \\
& 2 & 4.85 / 955 & \textbf{5.05} / \textbf{976} & 4.93 / 993 & \textbf{5.04} / \textbf{1000} & 3.11 / \textbf{664} & \textbf{3.12} / 653 \\
& 4 & 4.87 / 1718 & \textbf{5.11} / \textbf{1766} & 4.91 / 1796 & \textbf{5.00} / \textbf{1806} & 3.09 / 1236 & \textbf{3.16} / \textbf{1244} \\
& 8 & 4.91 / 2744 & \textbf{5.13} / \textbf{2822} & 4.95 / 2913 & \textbf{5.05} / \textbf{2928} & 3.12 / \textbf{2109} & \textbf{3.15} / 2099 \\
& 16 & 4.88 / 3961 & \textbf{5.07} / \textbf{4021} & 4.88 / 4175 & \textbf{4.94} / \textbf{4180} & 3.11 / \textbf{3312} & \textbf{3.12} / 3281 \\
& 32 & 4.86 / 4528 & \textbf{5.08} / \textbf{4649} & 4.85 / 4976 & \textbf{4.93} / \textbf{5008} & 3.09 / 4043 & \textbf{3.14} / \textbf{4102} \\
\addlinespace[4pt]
\multirow{6}{*}{\rotatebox[origin=c]{90}{\scriptsize \textit{$2K$}}}
& 1 & 5.45 / 560 & \textbf{5.67} / \textbf{573} & 5.16 / \textbf{541} & \textbf{5.25} / 541 & 3.12 / 339 & \textbf{3.18} / \textbf{339} \\
& 2 & 5.47 / 1025 & \textbf{5.71} / \textbf{1046} & 5.19 / \textbf{1001} & \textbf{5.25} / 997 & 3.15 / \textbf{649} & \textbf{3.19} / 648 \\
& 4 & 5.42 / 1751 & \textbf{5.68} / \textbf{1800} & 5.14 / \textbf{1724} & \textbf{5.22} / 1720 & 3.12 / 1180 & \textbf{3.17} / \textbf{1183} \\
& 8 & 5.48 / 2666 & \textbf{5.74} / \textbf{2713} & 5.20 / \textbf{2628} & \textbf{5.29} / 2623 & 3.16 / \textbf{1908} & \textbf{3.18} / 1907 \\
& 16 & 5.40 / 3550 & \textbf{5.68} / \textbf{3649} & 5.20 / 3594 & \textbf{5.32} / \textbf{3615} & 3.14 / 2870 & \textbf{3.18} / \textbf{2877} \\
& 32 & 5.41 / 4038 & \textbf{5.64} / \textbf{4091} & 5.18 / 4134 & \textbf{5.26} / \textbf{4179} & 3.14 / 3420 & \textbf{3.19} / \textbf{3441} \\
\bottomrule
\end{tabular*}}
\end{table}

\section{Loss-Term Ablation}
\label{app:ablation}

\chg{We train \method twice, with and without the distractor penalty of Eq.~\eqref{eq:distractor},
holding everything else fixed. The two arms share the architecture, every other setting, and the
total weight of the reranker objective, so removing the penalty hands its budget to the hard
cross-entropy rather than shrinking the objective. Table~\ref{tab:loss-ablation} gives the
per-benchmark acceptance lengths at concurrency one. The penalty raises acceptance length on all
nine benchmarks, by $1.5\%$ on the average, most on HumanEval at $3.8\%$ and least on AlpacaEval at
$0.2\%$. It is what trains the reranker against the candidates that actually cost acceptances
rather than against the ground-truth label alone, and Appendix~\ref{app:derivations} derives it as
an expected target cost.}

\begin{table}[t]
\centering
\footnotesize
\setlength{\tabcolsep}{4pt}
\caption{Loss-term ablation, Qwen3-8B target, greedy decoding, concurrency one, acceptance
length $\tau$. The two rows share the architecture, all other settings, and the total reranker
weight, so the second reallocates that budget rather than enlarging it. It is the model of
Section~\ref{sec:experiments}. Best per column in bold.}
\label{tab:loss-ablation}
\begin{tabular}{@{}l ccccccccc@{}}
\toprule
Objective & GSM8K & MATH & AIME & HEval & MBPP & LCB & Alpaca & MTB & SPEED \\
\midrule
$\mathcal{L}_{\mathrm{CE}} + \mathcal{L}_{\text{\dflash}}$
 & \chg{7.407} & \chg{9.031} & \chg{8.019} & \chg{6.819} & \chg{5.740} & \chg{7.579} & \chg{3.650} & \chg{3.923} & \chg{4.866} \\
$+\ \mathcal{L}_{\mathrm{dist}}$ (full)
 & \chg{\textbf{7.557}} & \chg{\textbf{9.064}} & \chg{\textbf{8.156}} & \chg{\textbf{7.077}} & \chg{\textbf{5.849}} & \chg{\textbf{7.754}} & \chg{\textbf{3.656}} & \chg{\textbf{3.939}} & \chg{\textbf{4.925}} \\
\bottomrule
\end{tabular}
\end{table}

\section{Derivations for the Reranker Objective}
\label{app:derivations}

\paragraph{The distractor penalty is an expected target cost.}
Write $P_\theta(k)$ for $P_\theta(c_s=k \given c_{s-1}=g_{s-1})$, let $c_k=-\log q_k$ be the
cost the target assigns to candidate $k$, low where the target is confident, and let
$w_k=\relu(c_k-c_{g_s})$ be that cost relative to the ground truth after rectification. The penalty
of Eq.~\eqref{eq:distractor} is then $\Omega_s=\sum_k P_\theta(k)\,w_k$, the expected rectified cost
of the candidate \method commits to. When $g_s$ is the target's preferred candidate in the pool no
cost is negative, the rectifier is inactive, and since $\sum_k P_\theta(k)=1$,
\begin{equation}
\Omega_s \;=\; \sum_k P_\theta(k)\,\left(c_k-c_{g_s}\right) \;=\; H(P_\theta,q)-c_{g_s},
\qquad
H(P_\theta,q)=\sum_k P_\theta(k)\,c_k .
\label{eq:omega-ce}
\end{equation}
$H(P_\theta,q)$ is the expected target cost under $P_\theta$, and the offset $c_{g_s}$ does not
depend on the \method parameters, so $\nabla_\theta\Omega_s=\nabla_\theta H(P_\theta,q)$: the
penalty and a cross-entropy against the target have the same gradient. The penalty involves the
target only through cost differences, so it is unchanged if $q$ is renormalized over the pool.

\paragraph{Why this is not distribution matching.}
The target is frozen, so the costs $c_k$ are constants and $\Omega_s$ is linear in $P_\theta$.
A linear function on the probability simplex attains its minimum at a vertex, that is, at a
distribution placing all of its mass on one candidate, and the minimizing vertex is the candidate
of least cost. Under the assumption above that candidate is $g_s$, so minimizing the penalty drives
\method to commit to the target's preferred candidate rather than to reproduce the target's
distribution over the pool. Committing is what the verifier rewards, since a block is accepted on
the tokens the target would itself emit and mass placed elsewhere is not recovered.

\paragraph{The rectifier.}
At the occasional slot where $g_s$ is not the target's preferred candidate, the rectifier
clips the negative costs that would otherwise reward a candidate the target ranks above the
supervised token. Such slots are rare, and arise from numerical differences between the offline
generation of the supervision and the teacher-forced target pass used during training, rather than
from a genuine target preference.

\paragraph{Complementary gradients.}
The hard cross-entropy and the distractor penalty act on the ground-truth logit $z_g$ through
complementary gradients. For cross-entropy,
\begin{equation}
\frac{\partial \mathcal{L}_{\mathrm{CE}}}{\partial z_g} = P_\theta(g) - 1,
\end{equation}
which remains strong when $P_\theta(g)$ is small, exactly the hardest mistakes. For the penalty,
\begin{equation}
\frac{\partial \Omega_s}{\partial z_g} = -\,P_\theta(g)\, \mathbb{E}_{P_\theta}[w],
\end{equation}
which vanishes as $P_\theta(g) \to 0$, so the penalty alone is weak precisely when the ground
truth is badly misranked. It is also indifferent among candidates of zero cost, including target
ties and candidates the target ranks above $g$, since the rectifier clips their costs to zero.
Cross-entropy therefore robustly pulls the exact ground-truth token upward, while $\Omega_s$
preferentially pushes target-rejected confusers downward, so the penalty is best viewed as
cost-aware shaping of cross-entropy rather than a replacement for it. Both terms are evaluated
only on the oracle prefix and averaged over the same slots, so slots after the first coverage miss
contribute to neither.

\section{Greedy Selection Versus the Highest-Scoring Path}

\label{app:decode}

Sections~\ref{sec:objective} and \ref{sec:decode} argue that prefix acceptance rewards a
greedy left-to-right decode over a search for the best-scoring block. We test the decode rule
in isolation, holding the trained model fixed and varying only the rule that selects a path
through the candidate lattice.

\paragraph{Path selection.}
Let $c=(c_1,\dots,c_{\block-1})$ denote a path, where $c_s$ is the candidate committed at
slot $s$ and $c_0$ is the anchor. Since \method couples only adjacent slots, a path scores as
the sum of its matching terms,
\begin{equation}
g(c) \;=\; \sum_{s=1}^{\block-1} \Epair\!\left(s\!-\!1,\, c_{s-1},\, c_s\right).
\label{eq:pathscore}
\end{equation}
The greedy rule of Section~\ref{sec:decode} commits each slot to the best continuation of
what precedes it and never revisits an earlier choice. The alternative is to return the path
that maximizes $g$. Enumeration is infeasible, as the lattice holds
$\topk^{\block-1}$ paths, but the pairwise structure of Eq.~\eqref{eq:pathscore} makes the
exact maximizer reachable in $O(\block\,\topk^{2})$ time by dynamic programming. Sweeping
left to right, one records for each candidate the score of the best path ending there,
then traces the stored predecessors back from the best final candidate. This is the Viterbi
algorithm \citep{viterbi1967}, a strictly stronger optimizer of $g$ than the greedy walk.

\paragraph{Why the stronger optimizer is \chg{not the better one here}.}
Under prefix acceptance every slot contributes only while all the slots before it have
matched, so the early ones carry most of the expected acceptance length, and a block that
loses its first slot gains nothing from the rest. Maximizing $g$ targets the block as a whole,
which is essentially the probability that the entire block survives verification. Giving up an
early slot to raise the total is rational for that objective and self-defeating here, because
the improved tail is never reached. The greedy walk instead maximizes the leading decision
exactly, and each later one conditioned on what was committed.

\begin{table}[t]
\centering
\small
\caption{Decode-rule ablation, Qwen3-8B target, greedy decoding, concurrency one, acceptance
length $\tau$. Both arms serve the same checkpoint and differ only in the rule that selects a path
through the lattice, each as a single fused kernel in the same captured graph, and the two run as
consecutive cells of one allocation. $\Delta$ is the
change from greedy to exact, so negative favors greedy. Best per row in bold.}
\label{tab:decode-ablation}
\begin{tabular}{lccc}
\toprule
Benchmark & Greedy & Exact & $\Delta$ \\
\midrule
SPEED             & \chg{\textbf{4.925}} & \chg{4.756} & \chg{$-3.43\%$} \\
GSM8K             & \chg{\textbf{7.557}} & \chg{7.408} & \chg{$-1.97\%$} \\
MATH-500          & \chg{\textbf{9.064}} & \chg{8.846} & \chg{$-2.40\%$} \\
AIME 2025         & \chg{\textbf{8.156}} & \chg{7.792} & \chg{$-4.46\%$} \\
HumanEval         & \chg{\textbf{7.077}} & \chg{7.018} & \chg{$-0.84\%$} \\
MBPP              & \chg{\textbf{5.849}} & \chg{5.714} & \chg{$-2.30\%$} \\
LiveCodeBench     & \chg{\textbf{7.754}} & \chg{7.633} & \chg{$-1.56\%$} \\
Alpaca            & \chg{\textbf{3.656}} & \chg{3.531} & \chg{$-3.42\%$} \\
MT-Bench          & \chg{\textbf{3.939}} & \chg{3.808} & \chg{$-3.32\%$} \\
\midrule
Mean $\Delta$     &       &       & \chg{$\mathbf{-2.64\%}$} \\
\bottomrule
\end{tabular}
\end{table}

\paragraph{Results.}
\chg{The comparison is between decode rules for one trained model, not between formulations:
\method is trained under the local objective of Eq.~\eqref{eq:localar}, so its path score is
calibrated for the walk it was trained to take. What Table~\ref{tab:decode-ablation} shows is that
this calibration is worth more than the stronger optimizer, at equal serving cost.}
\chg{The exact optimizer decodes worse.} It lowers acceptance
length on all nine benchmarks, \chg{by $0.84$ to $4.46\%$ and by $2.64\%$ on average}. Throughput follows,
since the two rules are equally cheap to serve: each is a single fused kernel folded into the same
captured draft graph, and \chg{the two rules run as consecutive cells of one allocation, where the
block rate they sustain differs by at most $0.44\%$ on any benchmark and by $0.06\%$ on average}. The stronger optimizer is therefore had for free and still \chg{accepts less}. Its
worst case is AIME 2025, though at $30$ problems that split carries the direction rather than the
magnitude; \chg{every other benchmark lies between $-0.84\%$ and $-3.43\%$. The smallest effect is on
HumanEval, which is also the smallest split and carries the largest block-rate excursion of the
nine.}

\section{Input-Length Sweep on the SPEED Throughput Split}
\label{app:isl}

Table~\ref{tab:isl-short} gives the acceptance lengths and absolute throughputs behind
Table~\ref{tab:concurrency}. Each block holds $512$
requests at a fixed input length with $1{,}024$ generated tokens, in each of the three output-entropy
tiers, and is swept from $c\!=\!1$ to $c\!=\!32$.

\chg{On the $1$K and $2$K blocks \method leads on both axes at every one of the thirty-six
operating points. Acceptance is ahead of the strongest system in each cell by
$1.1$ to $5.7\%$, throughput by $1.9$ to $8.2\%$, and no cell falls inside the tie band.}

\begin{table}[t]
\centering
\footnotesize
\setlength{\tabcolsep}{2pt}
\caption{\chg{Acceptance length and throughput behind Table~\ref{tab:concurrency}. Each cell holds $512$
requests with $1{,}024$ generated tokens. Reported as in Table~\ref{tab:main-8b}.}}
\label{tab:isl-short}
\begin{tabular*}{\textwidth}{@{}c@{\hspace{3pt}}c@{\extracolsep{\fill}}cccccc@{}}
\toprule
& $c$ & AR & \method & \dflashtwo & Domino & DSpark & \dflash \\
\midrule
\multirow{6}{*}{\rotatebox[origin=c]{90}{\scriptsize \textit{$1K$, low}}}
& 1 & 148 & \textbf{4.89} / \textbf{515} & 4.53 / 482 & 4.51 / 463 & 4.69 / 472 & 4.21 / 460 \\
& 2 & 283 & \textbf{4.85} / \textbf{955} & 4.50 / 898 & 4.50 / 863 & 4.69 / 879 & 4.20 / 863 \\
& 4 & 541 & \textbf{4.87} / \textbf{1718} & 4.55 / 1637 & 4.50 / 1559 & 4.69 / 1588 & 4.19 / 1558 \\
& 8 & 983 & \textbf{4.91} / \textbf{2744} & 4.54 / 2608 & 4.49 / 2479 & 4.72 / 2542 & 4.20 / 2493 \\
& 16 & 1705 & \textbf{4.88} / \textbf{3961} & 4.48 / 3730 & 4.50 / 3686 & 4.70 / 3725 & 4.19 / 3658 \\
& 32 & 2694 & \textbf{4.86} / \textbf{4528} & 4.53 / 4338 & 4.50 / 4319 & 4.69 / 4355 & 4.19 / 4212 \\
\addlinespace[4pt]
\multirow{6}{*}{\rotatebox[origin=c]{90}{\scriptsize \textit{$1K$, mixed}}}
& 1 & 148 & \textbf{4.91} / \textbf{526} & 4.66 / 502 & 4.74 / 491 & 4.81 / 490 & 4.25 / 471 \\
& 2 & 283 & \textbf{4.93} / \textbf{993} & 4.67 / 949 & 4.74 / 928 & 4.81 / 918 & 4.27 / 894 \\
& 4 & 543 & \textbf{4.91} / \textbf{1796} & 4.71 / 1742 & 4.79 / 1705 & 4.82 / 1688 & 4.26 / 1631 \\
& 8 & 986 & \textbf{4.95} / \textbf{2913} & 4.66 / 2772 & 4.79 / 2764 & 4.84 / 2716 & 4.27 / 2651 \\
& 16 & 1709 & \textbf{4.88} / \textbf{4175} & 4.67 / 4091 & 4.73 / 4038 & 4.83 / 3974 & 4.29 / 3900 \\
& 32 & 2745 & \textbf{4.85} / \textbf{4976} & 4.61 / 4844 & 4.69 / 4870 & 4.76 / 4834 & 4.21 / 4634 \\
\addlinespace[4pt]
\multirow{6}{*}{\rotatebox[origin=c]{90}{\scriptsize \textit{$1K$, high}}}
& 1 & 148 & \textbf{3.11} / \textbf{343} & 2.90 / 321 & 3.03 / 323 & 3.02 / 316 & 2.78 / 316 \\
& 2 & 285 & \textbf{3.11} / \textbf{664} & 2.89 / 617 & 3.05 / 628 & 3.02 / 606 & 2.77 / 606 \\
& 4 & 551 & \textbf{3.09} / \textbf{1236} & 2.92 / 1174 & 3.01 / 1161 & 3.00 / 1135 & 2.77 / 1148 \\
& 8 & 1016 & \textbf{3.12} / \textbf{2109} & 2.89 / 1965 & 3.03 / 1964 & 3.04 / 1943 & 2.80 / 1958 \\
& 16 & 1828 & \textbf{3.11} / \textbf{3312} & 2.89 / 3093 & 3.03 / 3172 & 3.01 / 3057 & 2.76 / 3061 \\
& 32 & 3041 & \textbf{3.09} / \textbf{4043} & 2.89 / 3877 & 3.01 / 3956 & 2.99 / 3846 & 2.77 / 3788 \\
\addlinespace[4pt]
\multirow{6}{*}{\rotatebox[origin=c]{90}{\scriptsize \textit{$2K$, low}}}
& 1 & 146 & \textbf{5.45} / \textbf{560} & 4.98 / 518 & 4.94 / 495 & 5.17 / 509 & 4.61 / 493 \\
& 2 & 276 & \textbf{5.47} / \textbf{1025} & 5.00 / 952 & 4.98 / 915 & 5.17 / 928 & 4.62 / 907 \\
& 4 & 514 & \textbf{5.42} / \textbf{1751} & 4.95 / 1644 & 4.95 / 1587 & 5.17 / 1621 & 4.58 / 1573 \\
& 8 & 898 & \textbf{5.48} / \textbf{2666} & 4.98 / 2487 & 4.93 / 2402 & 5.20 / 2463 & 4.59 / 2393 \\
& 16 & 1486 & \textbf{5.40} / \textbf{3550} & 5.02 / 3449 & 4.90 / 3321 & 5.14 / 3371 & 4.60 / 3304 \\
& 32 & 2206 & \textbf{5.41} / \textbf{4038} & 4.96 / 3862 & 4.92 / 3826 & 5.14 / 3864 & 4.58 / 3749 \\
\addlinespace[4pt]
\multirow{6}{*}{\rotatebox[origin=c]{90}{\scriptsize \textit{$2K$, mixed}}}
& 1 & 147 & \textbf{5.16} / \textbf{541} & 4.90 / 517 & 4.93 / 502 & 5.00 / 501 & 4.46 / 485 \\
& 2 & 278 & \textbf{5.19} / \textbf{1001} & 4.89 / 951 & 4.98 / 932 & 5.00 / 922 & 4.48 / 899 \\
& 4 & 520 & \textbf{5.14} / \textbf{1724} & 4.89 / 1665 & 4.92 / 1611 & 5.02 / 1610 & 4.49 / 1579 \\
& 8 & 907 & \textbf{5.20} / \textbf{2628} & 4.89 / 2520 & 4.97 / 2494 & 5.00 / 2439 & 4.48 / 2404 \\
& 16 & 1503 & \textbf{5.20} / \textbf{3594} & 4.94 / 3488 & 4.96 / 3466 & 5.00 / 3388 & 4.49 / 3322 \\
& 32 & 2267 & \textbf{5.18} / \textbf{4134} & 4.92 / 4026 & 4.97 / 4056 & 5.01 / 4017 & 4.49 / 3878 \\
\addlinespace[4pt]
\multirow{6}{*}{\rotatebox[origin=c]{90}{\scriptsize \textit{$2K$, high}}}
& 1 & 147 & \textbf{3.12} / \textbf{339} & 2.92 / 318 & 3.04 / 319 & 3.03 / 313 & 2.80 / 313 \\
& 2 & 279 & \textbf{3.15} / \textbf{649} & 2.95 / 612 & 3.05 / 611 & 3.05 / 594 & 2.82 / 604 \\
& 4 & 527 & \textbf{3.12} / \textbf{1180} & 2.93 / 1116 & 3.04 / 1113 & 3.03 / 1090 & 2.83 / 1108 \\
& 8 & 936 & \textbf{3.16} / \textbf{1908} & 2.92 / 1804 & 3.05 / 1810 & 3.05 / 1759 & 2.83 / 1786 \\
& 16 & 1591 & \textbf{3.14} / \textbf{2870} & 2.93 / 2727 & 3.04 / 2753 & 3.06 / 2694 & 2.76 / 2663 \\
& 32 & 2461 & \textbf{3.14} / \textbf{3420} & 2.94 / 3264 & 3.05 / 3338 & 3.04 / 3290 & 2.81 / 3205 \\
\bottomrule
\end{tabular*}
\end{table}

\end{document}